\def\year{2022}\relax
\documentclass[letterpaper]{article} 
\usepackage{aaai22}  
\usepackage{times}  
\usepackage{helvet}  
\usepackage{courier}  
\usepackage[hyphens]{url}  
\usepackage{graphicx} 
\usepackage{newfloat}
\usepackage{listings}
\usepackage{amssymb}
\usepackage{fontawesome5} 
\usepackage[most]{tcolorbox}
\usepackage{lipsum} 
\usepackage{natbib}  
\usepackage{caption} 
\DeclareCaptionStyle{ruled}{labelfont=normalfont,labelsep=colon,strut=off} 
\usepackage{algorithm}
\usepackage{algorithmic}
\usepackage{amsmath}
\usepackage{tabularray}
\usepackage{newfloat}
\usepackage{listings}
\floatstyle{ruled}
\newfloat{listing}{tb}{lst}{}
\floatname{listing}{Listing}

\title{DeepRule: An Integrated Framework for Automated Business Rule Generation via Deep Predictive Modeling and Hybrid Search Optimization}
\author {
    Yusen Wu,\textsuperscript{\rm 1}
    Xiaotie Deng, \textsuperscript{\rm 1}
}
\affiliations {
    \textsuperscript{\rm 1} School of Computer Science, Peking University\\
}

\usepackage{bibentry}

\begin{document}

\maketitle

\begin{abstract}
  This paper proposes DeepRule, an integrated framework for automated business rule generation in retail assortment and pricing optimization. Addressing the systematic misalignment between existing theoretical models and real-world economic complexities, we identify three critical gaps: (1) data modality mismatch where unstructured textual sources (e.g. negotiation records, approval documents) impede accurate customer profiling; (2) dynamic feature entanglement challenges in modeling nonlinear price elasticity and time-varying attributes; (3) operational infeasibility caused by multi-tier business constraints.
  Our framework introduces a tri-level architecture for above challenges. We design a hybrid knowledge fusion engine employing large language models (LLMs) for deep semantic parsing of unstructured text, transforming distributor agreements and sales assessments into structured features while integrating managerial expertise. Then a game-theoretic constrained optimization mechanism is employed to dynamically reconcile supply chain interests through bilateral utility functions, encoding manufacturer-distributor profit redistribution as endogenous objectives under hierarchical constraints. Finally an interpretable decision distillation interface leveraging LLM-guided symbolic regression to find and optimize pricing strategies and auditable business rules embeds economic priors (e.g. non-negative elasticity) as hard constraints during mathematical expression search. We validate the framework in real retail environments achieving higher profits versus systematic B2C baselines while ensuring operational feasibility. This establishes a close-loop pipeline unifying unstructured knowledge injection, multi-agent optimization, and interpretable strategy synthesis for real economic intelligence.
\end{abstract}

\section{Introduction}

Assortment selection and pricing are pivotal in market economies. This problem involves three interconnected components: First, modeling customer behavior by extracting multi-dimensional features from interactions to build personalized demand models. Second, decoupling product features to represent massive SKUs by separating intrinsic attributes from promotions. Third, constrained decision optimization to maximize revenue under business rules (e.g. inventory constraints, category requirements). Existing research shows significant progress across these dimensions.

\textbf{Customer Behavior Modeling: From Static Preferences to Dynamic Learning.} 
Accurate demand characterization underpins pricing. Traditional parametric models (e.g., MNL) face model misspecification and data sparsity challenges. For dynamic learning, \citet{context-based} proposes online clustering using generalized linear models with MLE, mitigating cold-start issues. Its algorithm dynamically constructs exploration via price perturbation, achieving $\tilde{O}(\sqrt{mT})$ regret. \citet{Bayesian-updates} integrates online reviews via Bayesian updates in bandit frameworks. For scenarios with high-dimensional features, \citet{high-dimensions} uses Regularized Maximum Likelihood Pricing (RMLP) with $O(s_0 \log d \cdot \log T)$ regret under sparsity.

\textbf{Product Feature Decoupling: Value Separation and Structural Compression.}
Core challenges lie in decoupling intrinsic value from promotions while handling high-dimensional features. \citet{ecommerce} designs a three-tier architecture: predicting baseline demand, computing elasticities for price-demand pairs, and solving for global prices via linear programming. This separates brand competition from elasticity differences. For compression, \citet{lowrank} proposes online SVD to learn low-rank subspaces, making OPOK/OPOL regret independent of product count. \citet{multi-product}'s M3P algorithm handles heterogeneity via phased exploration-exploitation with $O(\log(Td)(\sqrt{T} + d\log T))$ regret. Latent structure discovery remains key for dimensionality challenges.

\textbf{Constrained Decision Optimization: From Theoretical  Bou-nds to Industrial Deployment.}
The final decision stage must balance revenue maximization with business constraints. Current research advances focus on algorithmic design and engineering. For dynamic strategy frameworks in combinatorial optimization, \citet{assortmentselectionpricing} proposes the C-MNL model with a joint LCB-UCB strategy to achieve sublinear regret. \citet{mnldemand} extends this with optimistic algorithms for contextual bandits. In model-free optimization, \citet{model-free} constructs an Incentive-Compatible Polyhedron to enable robust pricing. For complex constraint handling in resource-constrained problems, \citet{primal-dual} introduces a Primal-Dual Dynamic Pricing algorithm with sublinear convergence. \citet{constrained-pricing} develops piecewise linear approximations and MILP solvers for constraints, extending to dynamic programming frameworks. In industrial-grade optimization engines, \citet{tricks} designs heuristics for cut generation to accelerate large-scale pricing, while \citet{model-distillation} proposes a "Prescriptive Student Tree" to distill black-box models into interpretable strategies.

\textbf{Emerging Scenarios and Open Challenges.}
Novel technologies like generative AI are catalyzing new pricing problems. For multi-task competition scenarios, \citet{pricing-competition} constructs a price-performance-ratio optimization framework with closed-form solutions. \citet{Perishability} formalizes data perishability challenges and proposes a Projected Stochastic Gradient Descent strategy achieving bounded regret. Regarding theoretical boundaries, \citet{towards-agnostic} and \citet{logarithmic-regret} establish fundamental minimax bounds and logarithmic regret results for algorithm design guidance. Despite significant algorithmic advancements, a systematic misalignment persists between theoretical frameworks and real-world economic complexities. Research predominantly focuses on systematized enterprises, neglecting the traditional retail sector with incomplete digitalization. In these settings, disparities exist between experimental and operational environments. Limited digital infrastructure constrains solution applicability, while conventional methods address isolated problems without full-chain integration. Experience-based rule algorithms fail to adapt to personalized demands, yielding suboptimal solutions with reliability issues. We identify four critical gaps:

\textbf{Data Modality Mismatch Impedes Customer Profiling.}
Existing methods predominantly rely on structured transactional data: high-dimensional pricing studies (e.g. \cite{high-dimensions}) require binary purchase decision feedback, while model-free approaches (e.g. \cite{model-free}) demand complete historical price-choice pairs. However, in traditional retail, critical decision inputs (e.g., distributor negotiation records, sales plan assessments) primarily exist as unstructured text. This absence renders parametric models (e.g., contextual MNL demand models in \cite{mnldemand}) ineffective. Although online clustering methods (e.g. \cite{context-based}) mitigate data sparsity through product grouping, their dependence on historical price-sales pairs prevents textual source processing, limiting precision in capturing customer heterogeneity.

\textbf{Dynamic Feature Entanglement Challenges Modeling Para-digms.}
When pricing large-scale SKUs (e.g. \cite{ecommerce}), disentangling multidimensional attribute cross-effects on demand is essential. While low-rank bandit methods (e.g. \cite{lowrank}) enhance scalability through feature subspace compression, their linear latent factor assumptions fail to capture abrupt elasticity shifts. Concurrently, robust optimization studies (e.g. \cite{random-robust}) address model uncertainty via randomized strategies but neglect feature time-variation (e.g. \cite{Perishability}), causing static feature models to degrade in long-term decisions.

\textbf{Operational Infeasibility from Multi-Tier Business Constraints.}
Real-world decisions are constrained by hierarchical business rules (e.g., \cite{primal-dual}) and regulations. Although constrained pricing research (e.g. \cite{constrained-pricing}) employs optimization for price boundaries, its approximations lack flexibility in handling localized policy elasticity. Crucially, industrial-scale solutions (e.g. \cite{tricks}) rely on historical solution reuse, inducing strategy degradation risks when frequent market-responsive adjustments are required.

\textbf{Interpretability Deficits Hinder Business Alignment.}
Black-box optimization models (e.g. Q-learning frameworks in \cite{q-learning}) adaptively adjust prices but suffer from decision opacity, preventing operational validation. While model distillation techniques (e.g. \cite{model-distillation}) extract readable rules, they disregard profit allocation requirements—such as multilayer agency problems (e.g. \cite{pricing-competition})—resulting in low adoption for cross-organizational coordination.

To systematically overcome these limitations, we propose DeepRule, an integrated decision framework that reconstructs assortment pricing through tri-level coordination:

\begin{enumerate}
    \item \textbf{Hybrid Knowledge Fusion Engine}: Addresses data fragmentation via deep semantic parsing, transforming unstructured text into structured feature representations while integrating expert-defined rules to create joint supervision signals. This resolves data sparsity and business knowledge fragmentation.
    
    \item \textbf{Game-Theoretic Constrained Optimization}: Tackles objective conflicts by jointly modeling pricing and sales volume dynamics within a unified framework. It incorporates feature evolution patterns and business constraint networks, introducing bilateral utility functions that transform manufacturer-distributor profit redistribution into endogenous optimization objectives.
    
    \item \textbf{Interpretable Decision Distillation Interface}: Ensures operational feasibility by translating complex strategies into auditable business logic via rule extraction, with a distributed optimization architecture enabling efficient solving of massive decision variables.
\end{enumerate}

The core innovation establishes the first close-loop decision flow integrating \textit{unstructured knowledge injection}, \textit{dynamic constrained game modeling}, and \textit{interpretable strategy generation}, maintaining both academic rigor and industrial robustness. Empirical validation demonstrates significant reduction of the optimality-operationality gap across traditional retail scenarios, providing a replicable engineering paradigm for AI-driven economic intelligence.

\section{Related Work}
Symbolic Regression (SR) autonomously discovers mathematical expressions from data without presupposing functional forms. Unlike conventional regression techniques, SR explores spaces of operators, variables, and constants to generate interpretable equations that unveil underlying patterns through selective rule-model iteration.

\subsection{Evolving of Symbolic Regression Methods}
SR is extensively employed in scientific discovery. Early approaches relied on Genetic Programming (GP) for expression space exploration, but faced limitations in diversity maintenance and computational efficiency \cite{sr-review}. Deep Symbolic Regression (DSR) addressed these by formulating expression generation as sequential decision-making, where \citet{deep-sr} prioritizes optimization of high-reward expression subsets to mitigate local optima. However, traditional DSR suffers from limited long-range dependency modeling. End-to-end Transformers leverage self-attention to globally model variable relationships, with \citet{symformer} pioneering encoder-decoder architectures that eliminate iterative search overhead. To address symbol-coefficient coupling, \citet{transformer-plan} proposes dual-objective optimization combining supervised data fitting and contrastive learning. Large Language Models (LLMs) fuse prior knowledge with few-shot reasoning, where In-Context Learning \citet{sr-llm-discovery} demonstrates rule inference from minimal examples. For mathematical reasoning enhancement, \citet{gpt-guided} introduces GPT-guided Monte Carlo Tree Search (MCTS) for directed expression space exploration, validating hybrid architectures combining LLM generation with search refinement.

\subsection{Domain-Guided SR for Economic Constraints}
We integrate SR with physical economy business rules for interpretable optimization. Economic pricing requires adherence to domain constraints (e.g. diminishing marginal returns), necessitating domain-guided SR. \citet{sr-physical} incorporates dimensional consistency as hard constraints for physically valid expressions, extendable to economic priors. SR's interpretability advantage over "black-box" models lies in concise mathematical expressions that explicitly reveal factor relationships, enabling auditing \cite{sr-review}.

\section{Preliminary}
Building on reviewed literature and aligning with our optimization objectives for four business challenges, we systematically reproduce and compare four influential baseline algorithms in their respective methods. These algorithms address the defined real-world economic challenges with unique contributions but exhibit specific limitations.

\subsection{Low-Rank Bandit Methods for High-Dimensional Dynamic Pricing}
\citet{lowrank} leverages low-rank structural properties of demand models to overcome dimensionality in multi-product pricing. The core assumption posits that cross-product demand elasticities are captured by a $d$-dimensional latent feature matrix $\mathbf{U} \in \mathbf{R}^{N \times d}$ ($d \ll N$), formalized through:
\[
\mathbf{q}_t = \mathbf{U}\mathbf{z}_t - \mathbf{U}\mathbf{V}_t\mathbf{U}^T\mathbf{p}_t + \boldsymbol{\epsilon}_t
\]
Optimization in projected space $\mathbf{x} = \mathbf{U}^T\mathbf{p}$ yields quadratic convex revenue $f_t(\mathbf{x}) = \mathbf{x}^T\mathbf{V}_t\mathbf{x} - \mathbf{x}^T\mathbf{z}_t$. The method achieves $\mathcal{O}(d\sqrt{T})$ regret independent of $N$, offering superior scalability. However, performance critically relies on presumed linear low-rank structure, leading to suboptimality under non-low-rank scenarios or high noise.

\subsection{Context-Based Dynamic Pricing with Online Clustering}
Designed for data-sparse products, \citet{context-based} enables cross-product data sharing via real-time clustering of demand patterns. It operates on a generalized linear model:
\[
\mathbf{E}[d_{i_t,t}] = \mu(\alpha_{i_t}^\top x_t + \beta_{i_t} p_{i_t,t})
\]
The algorithm constructs neighborhoods $\hat{\mathcal{N}}_{i,t}$ of products with similar demand parameters, computing semi-myopic cluster-level prices $p'_{i,t}$. Exploration uses perturbations $\Delta_{i,t} = \pm \Delta_0 \widetilde{T}_{\hat{\mathcal{N}}_{i,t},t}^{-1/4}$, improving single-product pricing. Limitations include sensitivity to feature distributions (high dimensions yield poor clusters) and restricted generalization under distributor behavior fluctuations.

\subsection{Model-Free Assortment Pricing with Transaction Data}
\citet{model-free} circumvents parametric assumptions via incentive compatibility (IC) constraints. Customer $i$'s behavior defines a valuation polyhedron:
\[
V_i = \left\{ \mathbf{v}_i : v_{ic_i} - P_{ic_i} \geq \max_{j' \neq c_i}(v_{ij'} - P_{ij'}) \geq 0 \right\}
\]
Optimal pricing uses minimax robust optimization:
\[
\tau^* = \sup_{\mathbf{p} \geq 0} \frac{1}{m} \sum_{i=1}^m \min_{\mathbf{v}_i \in V_i} \left[ \max_j p_j \cdot \mathbf{I}_{\text{rational choice}} \right]
\]
The approach provides robustness against model misspecification but loses efficacy in high-dimensional spaces with smooth gradients and personalization-intensive scenarios, while exploration capacity is limited in constrained decision spaces.

\subsection{B2B Product Offers with Machine Learning, Mixed Logit, and Nonlinear Programming}
Combining Mixed Logit Models (MLM) with Nonlinear Programming (NLP), \citet{b2b} segments customers via Random Utility Theory. Customer $c$'s utility for offer $j$ is:
\[
U_{cj} = k_j + \boldsymbol{x}_{cj}\boldsymbol{\beta}_c + \epsilon_{cj}
\]
with choice probability $\text{Prob}_{cj} = {e^{U_{cj}}}/{\sum_i e^{U_{ci}}}$. Optimal segment-level pricing $(r_{e,l}, M_{e,l})$ maximizes:
\[
\max_{r,M} \sum_c \text{Prob}_{c,o} \times \text{Loyalty}_c \times \sum_{m=1}^{M} \frac{\text{Net Revenue}_m}{(1+d/12)^m}
\]
The structured model provides interpretability but heavily depends on business priors and manual classifications without automated updates, necessitating resource-intensive market adaptations.

\section{Methods}
Here we make description about our core pipeline, while formulated details can be found in appendix. 

\begin{figure}
  \includegraphics[width=0.48\textwidth]{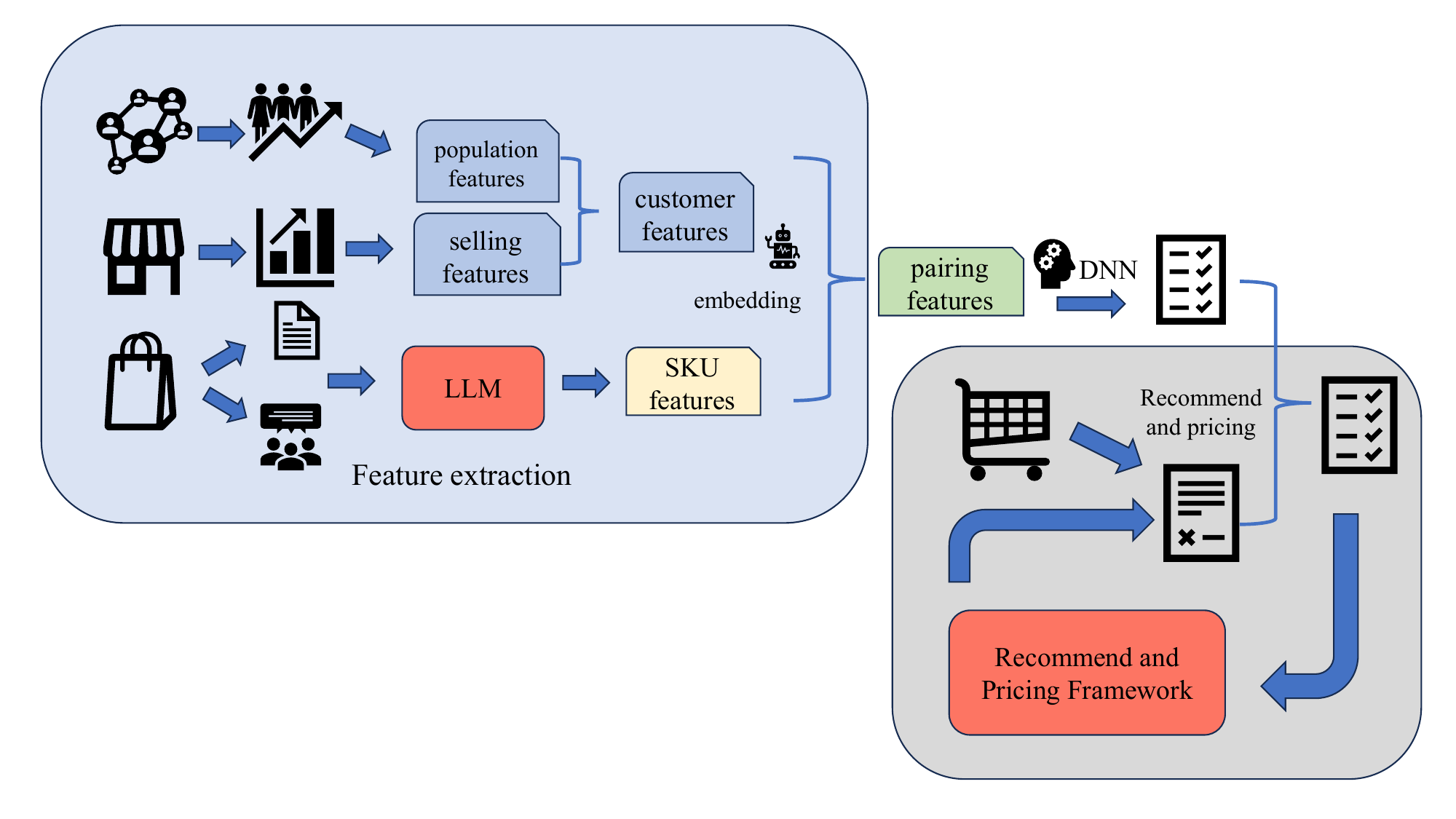}
  \caption{DeepRule Framework. }
  \label{fig:framework}
\end{figure}

\subsection{Sales Volume-Investment Depth Prediction Model}

\subsubsection{Feature Extraction and Data Curation}

\textbf{Mathematical Modeling Framework.} 
To address partial observability of store-customer affiliations, we leverage LLM technology to extract key prior knowledge from multi-source data, construct customer features, and refine them via feature engineering. The pipeline progresses from data parsing to feature aggregation, yielding efficient representations. Inputs include customer public information, historical records, price applications, approval documents, geolocation coordinates, demographics, and SKU attributes. Outputs are standardized customer feature vectors $\mathbf{f}_k \in \mathbf{R}^{d_f}$ and SKU encodings $\mathbf{e}_{\text{SKU}} \in \mathbf{R}^{d_s}$ for downstream tasks.

\textbf{LLM Analysis and Prior Knowledge Extraction.} 
A locally deployed LLM parses unstructured public information to generate store affiliation priors, reward functions, and decision basis sets. For business scope and brand preferences, semantic parsing extracts entities and relations, outputting structured prior vectors $\mathbf{p}_{\text{prior}} \in \mathbf{R}^d$. For reward/penalty records, a sequence-to-sequence framework transforms text sequences into reward functions $R(s, a)$, formalized as $R(s, a) = \text{MLP}(\mathbf{h}_{\text{cls}})$. For price applications and approvals, LLM analysis associates numerical features with text sentiment to construct decision sets $\mathcal{D} = \{ (\mathbf{x}_i, y_i) \}$. The unified outputs—$\mathbf{p}_{\text{prior}}$, $R(s,a)$, and $\mathcal{D}$—provide foundational knowledge for feature aggregation.

\textbf{Multi-dimensional Demographic Profile Aggregation Strategy.} 
To address affiliation partial observability, we design a weighted aggregation mechanism using spatial distance and operational assumptions. Spherical distance $d_{jk}$ between store $m_j$ and customer $c_k$ is computed via the Haversine formula. Customer operational radius $r_k$ correlates with business scale as $r_k = \alpha s_k$. Demographic features $\mathbf{g}_k$ aggregate profiles of stores within $c_k$'s operational radius. For stores covered by multiple customers, affiliation priority uses LLM priors: if similarity exceeds threshold $\theta$, $m_j$ affiliates exclusively with $c_k$. Otherwise, distance-scale weighting applies as $w_{jk} = \frac{s_k / d_{jk}^2}{\sum_{c \in \mathcal{C}_j} s_c / d_{jc}^2}$, yielding aggregated features $\mathbf{g}_k = \sum_{j \in \mathcal{M}_k} w_{jk} \mathbf{q}_j$. Temporal features augment via window statistics, Fourier coefficients, and sequential features. The final customer representation is $\mathbf{f}_k = [\mathbf{g}_k; \mathbf{stats}_k]$.

\textbf{Feature Standardization and Encoding.}
To address feature scale disparities, we apply Z-score standardization $x'_{i} = \frac{x_i - \mu_i}{\sigma_i}$. Dual-tower encoding prevents critical feature dominance: the high-dimensional tower processes concatenated features, the tower for low-dimension encodes key variables like sales volume, and weighted fusion yields $\mathbf{e}_{\text{final}} = [\beta \mathbf{e}_{\text{high}}; (1-\beta) \mathbf{e}_{\text{low}}]$, with $\beta < 0.5$ optimized via cross-validation.

\textbf{SKU Feature Representation.}
To mitigate sparsity in SKU attributes, style embeddings bypass one-hot encoding: text-based methods generate $\mathbf{e}_{\text{style}} = \text{PCA}(\text{Bert}(\mathbf{t}_{\text{style}}))$, while multimodal methods decompose attributes into orthogonal components projected as $\mathbf{e}_{\text{style}} = \mathbf{W}[\mathbf{c}; \mathbf{patt}; \mathbf{scent}] + \mathbf{b}$. Additional features are processed directly or embedded according to being sparse or dense, forming the unified representation $\mathbf{e}_{\text{SKU}} = [\mathbf{e}_{\text{style}}; \mathbf{e}_{\text{dense}}; \mathbf{e}_{\text{sparse}}]$.

This pipeline ensures logical transformation from raw data to feature representations: LLM-extracted priors drive profile aggregation, standardized features integrate with SKU encodings, and dual-tower encoding preserves critical information for downstream tasks. The mathematically defined transformations emphasize scalability and domain knowledge integration.

\subsubsection{Prediction with DNN model}

\textbf{Feature-Decoupled Sales Volume Modeling.}
To eliminate scale interference from pricing, we map multi-dimensional features to shipment units. Input feature tensor $\mathbf{X} = [\mathbf{X}_{\text{cust}} \oplus \mathbf{X}_{\text{sku}} \oplus \mathbf{X}_{\text{price}} \oplus \mathbf{X}_{\text{promo}}]$ is processed by a DNN $f_\theta$ modeling conditional expectation $\hat{y}_{\text{units},i} = f_\theta(\mathbf{x}_i)$. Revenue is decoupled explicitly through $\hat{s}_{\text{revenue},i} = \hat{y}_{\text{units},i} \cdot x_{\text{price},i}$. Parameters optimize by minimizing regularized mean squared error $\mathcal{L}_{\text{pred}} = \frac{1}{N} \sum_{i=1}^N \left[ \left( f_\theta(\mathbf{x}_i) - y_{\text{units},i} \right)^2 + \lambda \|\theta\|_F^2 \right]$, controlling complexity to prevent overfitting.

\textbf{Rule-Prior Guided Data Augmentation with Error Control.}
To address manual annotation constraints, an RLAIF\cite{rlaif}-inspired approach is adopted. Given a labeling rule $g: \mathcal{X} \to \{0,1\}$ with annotation noise $\delta \sim \text{subG}(\sigma^2)$, a pretrained LLM $\Phi_{\text{LLM}}$ generates pseudo-labels $\tilde{y}_j = \Phi_{\text{LLM}} \left( \mathcal{P}_{\text{prompt}}(g, \mathbf{x}_j) \right)$. Pseudo-label bias satisfies $\mathbf{E}[|\tilde{y}_j - y_{\text{true},j}|] \leq \eta_{\text{LLM}}$ with $\eta_{\text{LLM}} = k \cdot \exp(-I(g)/\tau)$. For augmented data $\mathcal{D}_{\text{aug}}$, the generalization error of $f_\theta$ admits $R_{\text{true}}(f_\theta) \leq \widehat{R}_{\text{aug}}(f_\theta) + \mathcal{C}(f_\theta, \delta) + \Gamma_{\text{noise}}$. When $I(g) > \tau \ln(k/\sigma)$, label noise remains at human-annotation level $\sigma$, and pseudo-labeling achieves comparable error control.

\textbf{Multi-Source Weighted Fusion for Prediction Optimization.}
The rule base $\mathcal{R} = \{\gamma_k\}_{k=1}^K$ is partitioned into strict constraints and soft recommendations. Relevant rules are retrieved via RAG using $\gamma_t^* = \underset{\gamma_k \in \mathcal{R}}{argmax} \cos(\psi_{\pi}(\mathbf{x}_t), \psi_{\pi}(\gamma_k))$. Dynamic weight $\alpha_t$ and strictness flag $h_t$ are generated via LLM. Initial prediction $a_{\text{init}} = f_\theta(\mathbf{x}_t)$ is fused with rule output: $a_{\text{fused}} = g_{\text{strict}}(\mathbf{x}_t)$ if $h_t = 1$, else $\alpha_t a_{\text{init}} + (1 - \alpha_t) g_{\text{soft}}(\mathbf{x}_t)$. Historical trend calibration applies differential adjustment $\Delta H_t = \frac{1}{\tau} \sum_{i=1}^\tau \left( a_{t-i} - a_{t-i-\tau} \right)$, yielding $a_{\text{final}} = \beta \cdot \text{sgn}(\Delta H_t) \cdot \min(|\Delta H_t|, \kappa) + (1 - \beta) a_{\text{fused}}$. Validated results update the rule base.

\textbf{Business Data Posterior Cleaning and Validation.}
After obtaining predictions $\hat{y}_i = a_{\text{final}}(\mathbf{x}_i)$ on training/confidence set $D_{\text{train}}$, low-confidence samples are identified using a confidence metric $\mathcal{C}(\mathbf{x}_i, \hat{y}_i, y_i)$ and threshold $\delta$. These undergo LLM-based cleaning:
\begin{equation}
(\text{judgment}_i, \text{reason}_i, \text{rule}_i) = \Phi_{\text{LLM}} \left( \mathcal{P}_{\text{clean}} \right)
\end{equation}
where prompt $\mathcal{P}_{\text{clean}}$ requests label validity assessment and cleaning rule generation. Samples judged invalid are corrected/removed, with generated rules updating the labeling repository. Ambiguous cases (judged "uncertain") undergo manual analysis. Human-verified outcomes refine the rule base and model, establishing a closed-loop data quality system.

\subsection{Assortment-Pricing Rule Search}

\begin{figure}
  \includegraphics[width=0.48\textwidth]{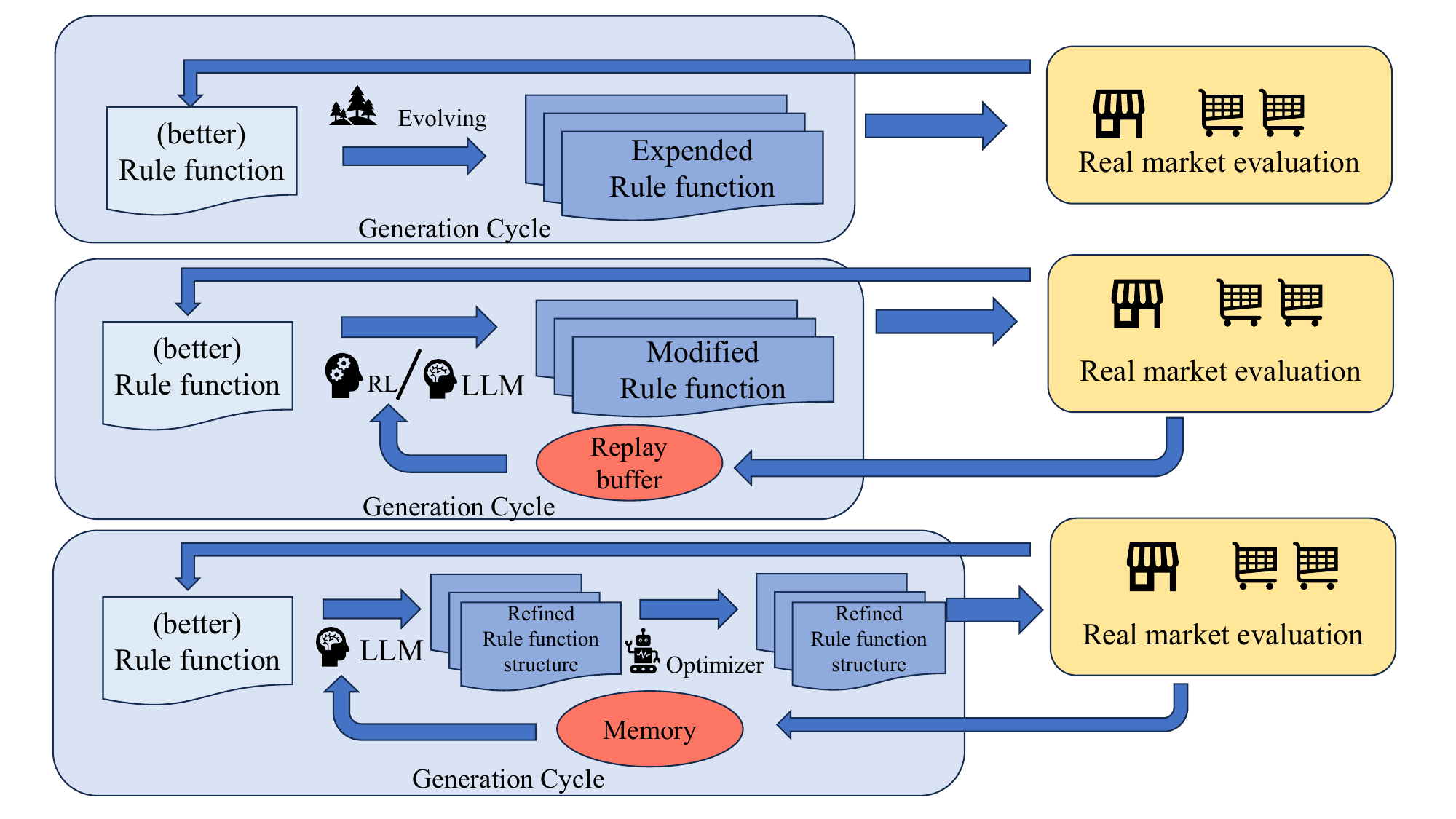}
  \caption{Assortment-pricing rule search methods. }
  \label{fig:policy}
\end{figure}

Following the fundamental principles of symbolic regression, we conduct a search for assortment-pricing strategies. Each strategy is formalized as a function string that simultaneously outputs a binary assortment decision (whether to recommend stocking a specific SKU for a customer) and a recommended price. We experimentally evaluate several rule search methodologies, including an evolutionary decision tree search algorithm, reinforcement learning (RL) algorithms, and a large language model (LLM)-based reasoning iteration algorithm.

\subsubsection{Symbolic Regression for Assortment-Pricing Strategy Modeling}
We model assortment-pricing decisions via an interpretable function $f: \mathbf{R}^n \to \{0,1\} \times \mathbf{R}^+$. For feature vector $\boldsymbol{x}$ (encoding market dynamics, inventory costs, etc.), the function outputs:
\[
(d, p) = 
\begin{cases} 
(1, g(\boldsymbol{x})) & \text{if } h(\boldsymbol{x}) > \tau \\
(0, \varnothing) & \text{otherwise}
\end{cases}
\]
where $h$ determines assortment, $g$ determines pricing, and $\tau$ is a threshold. Symbolic regression searches for $f^*$ in space $\mathcal{F}$ (arithmetic operators, exponents, etc.) via:
\[
f^* = \arg\max_{f \in \mathcal{F}} \Phi(f; \mathcal{D}_{\text{test}})
\]
with $\Phi$ evaluating performance on test data.

\subsubsection{Evolutionary Algorithm}
The evolutionary algorithm optimizes symbolic expressions through natural selection. An initial population $\mathcal{P}_0$ of $M$ expression trees is randomly generated. At generation $t$, generate candidates $\mathcal{C}_t$ by mutating $\mathcal{P}_t$ with Structural Expansion - Insert random subtree at random node, Structural Pruning - replace random subtree with leaf node, parameter Perturbation - adjust constants with Gaussian noise. Then we do fitness Evaluation on $\Phi(f)$ for $\mathcal{C}_t \cup \mathcal{P}_t$:
\begin{equation}
\Phi(f) = \frac{1}{|\mathcal{D}_{\text{test}}|} \sum_{i=1}^{N} \left[ \mathbf{I}_{d_i=1} (p_i - c_i) q_i(p_i) - \eta (p_i - p_i^*)^2 \right] - \lambda \ell(f)
\end{equation}
where Term 1 computes profit for stocked items, Term 2 penalizes market price deviation, and Term 3 penalizes complexity $\ell(f)$ (node-weighted sum). Finally make selection based on environment interaction. Merge $\mathcal{P}_t$ and $\mathcal{C}_t$, rank by $\Phi(f)$, and select top $M$ individuals for $\mathcal{P}_{t+1}$. Iterate above process until fitness converges or generation limits are reached.

\subsubsection{Reinforcement Learning Framework}
Symbolic expression generation is modeled as a Markov Decision Process (MDP). States $\mathcal{S}$ are partial expression trees, actions $\mathcal{A}$ involve node operations, and transitions are deterministic. The sparse reward is:
\[
r(s_t, a_t) = \Phi(f_{s_T}) \text{ at terminal state, } 0 \text{ otherwise}
\]
We optimize policy $\pi_\phi(a|s)$ via off-policy Q-learning. Experiences collected under $\epsilon$-greedy behavior populate replay buffer $\mathcal{B}$. The Q-function updates by minimizing:
\[
\mathcal{L}(\psi) = \mathbf{E}_{\mathcal{B}} \left[ \left( Q_\psi(s_t,a_t) - \hat{Q}(s_t,a_t) \right)^2 \right]
\]
with target $\hat{Q}(s_t,a_t) = r_t + \gamma \max_{a'} \bar{Q}_{\bar{\psi}}(s_{t+1}, a')$. Policy parameters $\phi$ update via gradient ascent. Target parameters $\bar{\psi}$ periodically synchronize with $\psi$, and $\pi^*$ generates expressions greedily.

\textbf{Unified Reward Function Design}
Both algorithms employ identical $\Phi(f)$ with three components: Expected Profit, $(p_i - c_i) \cdot q_i(p_i)$ using demand predicted via historical elasticity; Price Deviation Penalty, $\eta \cdot (p_i - p_i^*)^2$; Model Complexity Penalty, $\lambda \cdot \ell(f)$ via weighted node count. Coefficients $\eta$ and $\lambda$ balance objectives, determined via cross-validation.

\subsubsection{Generative Large Language Model (LLM) Based Rule Search Framework}
We formulate the assortment-pricing decision problem as finding an optimal function $f^*: \mathcal{X} \to \mathcal{Y}$, where $\mathcal{X} \subseteq \mathbf{R}^d$ represents customer feature vectors (e.g., purchasing power, category preference) and $\mathcal{Y}$ denotes assortment-pricing plans. The business evaluation function $L(f) = \mathbf{E}_{(x,y^*)\sim \mathcal{D}} \left[ \ell(f(x), y^*) \right] + \lambda \cdot \Omega(f)$ quantifies performance, where $\ell(\cdot)$ measures deviation from ideal plan $y^*$, $\Omega(f)$ penalizes rule complexity based on business priors, and $\lambda$ balances trade-offs. Traditional RL methods optimize $f_\theta$ via policy gradients but struggle with convergence in high-dimensional discrete action spaces. To address this, we deploy a generative LLM as meta-optimizer for symbolic regression, leveraging key advantages including Symbolic Rule Generation (Outputs human-interpretable mathematical expressions), Latent Space Guidance (Compresses search space via pre-trained semantic knowledge), and Reflection Mechanism (Generalizes error patterns through memory).

\subsubsection{Iterative Rule Optimization with LLMs}
Domain-expert business rules initialize the model. These rules, their validation results, and relevant prompts form first-round input to initiate the LLM's reflection phase.

The decision rule at iteration $t$ is $f_t(x) = \text{LLM}_{\text{reason}}(x; M_{<t})$, with memory bank $M_t = \{(c_k, \Delta f_k, \delta L_k)\}_{k=1}^K$ storing triples (customer type $c_k$, rule modification $\Delta f_k$, loss change $\delta L_k$). The optimization proceeds as:

\begin{enumerate}
    \item \textbf{Rule Generation:} Initialize via prompt template:
    \begin{equation}
    \begin{split}
        f_0(x) &= \text{LLM}(\mathcal{P}_{\text{init}}; \emptyset) \\
        \mathcal{P}_{\text{init}} &= \text{``Generate a mathematical expression for} \\
        &\text{the assortment-pricing function } f(x) \text{''}
    \end{split}
    \end{equation}
    
    \item \textbf{Rule Evaluation:} Compute segment-wise loss:
    \[
    L_t^{(c)} = \frac{1}{| \mathcal{D}_c |} \sum_{(x,y^*)\in \mathcal{D}_c} \ell(f_t(x), y^*),  \quad \forall c \in \mathcal{C}
    \]
    where $\mathcal{D}_c$ is the validation dataset for customer type $c$.
    
    \item \textbf{Reflection and Refinement:} Construct refinement prompt:
    \[
    \mathcal{P}_{\text{refine}} = \mathcal{P}_{\text{base}} \oplus \{(c, L_t^{(c)}, \nabla_f L_t^{(c)})\}_{c\in \mathcal{C}_{\text{fail}}} \oplus \text{TopK}(M_{<t}, \text{sim}(c,c'))
    \]
    where $\mathcal{C}_{\text{fail}} = \{c \mid L_t^{(c)} > \tau\}$ identifies high-loss segments, $\text{sim}(\cdot)$ computes embedding similarity, and $\oplus$ concatenates prompts. Generate refined rule:
    \[
    f_{t+1}(x) = \text{LLM}(\mathcal{P}_{\text{refine}}; \Theta)
    \]
    
    \item \textbf{Memory Update:} For modified customer type $c$:
    \[
    M_t \leftarrow M_t \cup \left\{ \left( c, \Delta f_t^{(c)}, L_t^{(c)} - L_{t-1}^{(c)} \right) \right\}
    \]
\end{enumerate}
Termination occurs when $\max_{c} L_t^{(c)} < \epsilon$ or $t \geq T_{\max}$. This approach transforms rule search into a sequential decision problem in semantic space, with convergence ensured by memory coverage of error patterns.

\subsubsection{Decoupled Function Structure Generation and Parameter Optimization}
To address combinatorial explosion, we propose hierarchical optimization. For function structure $f$ with $k$ parameters $\theta \in \Theta \subset \mathbf{R}^k$, direct discretization yields intractable search complexity. 

\begin{itemize}
    \item \textbf{Phase 1: Function Structure Generation} \\
    Generate constrained symbolic structures via LLM:
    \[
    f(x) = \sum_{j=1}^J \alpha_j h_j(x) + \sum_{p=1}^P \beta_p \cdot \mathbf{I}(x \in \mathcal{R}_p)
    \]
    where $h_j$ are basis functions (e.g., linear/cross terms, sigmoid), and $\mathcal{R}_p$ denote feature space regions.
    
    \item \textbf{Phase 2: Parameter Optimization} \\
    Fix structure and solve continuous optimization:
    \[
    \theta^* = argmin_{\theta \in \Theta} L(f_\theta)
    \]
    using gradient-based methods (e.g. Sequential Quadratic Programming, BFGS). If optimized loss $L(f_{\theta^*}) > \xi$, reconstruct the framework via LLM prompting.
\end{itemize}
This two-stage loop reduces complexity by decoupling combinatorial structure search and continuous parameter optimization.

\subsection{Example of Hierarchical Optimization Algorithm Rules}
Integrating the predictive model with our rule search framework yields novel algorithmic rules demonstrating superior simulated performance. The optimized decision function is semantically summarized as a comprehensive assortment-pricing strategy, generating procurement plans through a four-stage pipeline:
\begin{enumerate}
    \item Data preprocessing
    \item Sales forecasting
    \item Cost and fee-ratio calculation
    \item Constrained optimization
\end{enumerate}
The pipeline begins by cleansing and fusing multi-source data (sales, materials, customer, cost) for predictive modeling. Subsequent stages forecast sales, quantify unit costs and category-level fee ratios, then allocate material sales volumes under business constraints to maximize profit.

\subsubsection{Data Preprocessing}
This stage constructs a unified input dataset through customer filtering, metric computation, and data integration. Consider customer set $\mathcal{C}$ and material set $\mathcal{M}$. Historical sales volumes $s_{c,m}$ and unit prices $p_{c,m}$ are processed alongside forecasted sales $\hat{s}_c$. Candidate customers are filtered via thresholding:  
\[
\mathcal{C}_{\text{cand}} = \{ c \mid \hat{s}_{c} > \theta \}, \quad \theta = 1.0
\]
to target high-potential customers and reduce computation.  

Unit prices are directly extracted or computed from sales amounts. Material attributes (codes, categories) and customer identifiers are integrated through inner joins, forming a base dataset. Weighted unit prices $\bar{p}_m$ are computed as sales-weighted market averages. Cost data is matched to customers via string similarity, with the preprocessed output combining base data, weighted prices, and matched costs.

\subsubsection{Sales Forecasting and Data Augmentation}
This stage generates future sales predictions using enhanced data and ML models. For each $c \in \mathcal{C}_{\text{cand}}$:
\begin{enumerate}
    \item \textbf{Expand feature space}: Form customer-material pairs $\mathcal{P}_{c} = \{ (c, m) \mid m \in \mathcal{M} \}$ via Cartesian product
    \item \textbf{Simulate discount scenarios}: Apply discrete discount levels $d \in \mathcal{D}$ (e.g., $\{0.9, 0.95, 1.0\}$), constructing augmented dataset $\mathbf{D}_{\text{aug}}$ with feature vectors and target $s_{c,m}$
\end{enumerate}
Dataset size becomes $|\mathcal{C}_{\text{cand}}| \times |\mathcal{M}| \times |\mathcal{D}|$.  

A supervised model uses z-score normalized features and pretrained $f_{\text{model}}$ to predict sales:
\[
\hat{s}_{c,m,d} = f_{\text{model}}(\mathbf{x}_{\text{norm}})
\]
Baseline predictions $\hat{s}_{c,m} = \hat{s}_{c,m,d=1.0}$ feed downstream optimization.

\subsubsection{Cost and Fee Ratio Calculation}
This phase computes cost metrics and category-level fee ratios. Unit costs $c_{\text{cost},m}$ are sourced directly. Fee ratios (expense-to-revenue) use hierarchical allocation:  
\begin{enumerate}
    \item \textbf{Gross Revenue (GR)}: Historical $\text{GR}_{c,m} = s_{c,m} \cdot p_{c,m}$; category-level $\text{GR}_{c,k_1} = \sum_{m \in \mathcal{M}_{k_1}} \text{GR}_{c,m}$
    \item \textbf{Category proportion}: $h_{c,k_1} = \text{GR}_{c,k_1} / \sum_{k_1'} \text{GR}_{c,k_1'}$
    \item \textbf{Fee ratios}: For expense types $\mathcal{F}$, 
    \begin{equation}
        \begin{split}
            \text{FR}_{c,k_1,f_t} &= h_{c,k_1} \cdot \text{TotalExpense}_{f_t} / \text{GR}_{c,k_1}, \\
            \text{FR}_{c,k_1} &= \sum_{f_t \in \mathcal{F}} h_{c,k_1} \text{FR}_{c,k_1,f_t}
        \end{split}
    \end{equation}
\end{enumerate}
Monthly averaged GR $\overline{\text{GR}}_c$ constrains optimization. Final prediction data $\mathbf{D}_{\text{pred}}$ merges forecasts, categories, costs, and fee ratios.

\subsubsection{Constrained Optimization Decision}
This stage allocates recommended sales volumes under volatility and distribution constraints to maximize profit. Define decision variable $x_{c,m}$ as sales amount for material $m$ to customer $c$, with profit $\pi_{c,m} = x_{c,m} \cdot (p_{c,m} - c_{\text{cost},m})$. Optimization applies multi-tier constraints. In global, historical average order value $A_c$ bounds total sales $X_c = \sum_m x_{c,m} \cdot p_{c,m}$:
  \[
  0.95 A_c \leq X_c \leq 1.05 A_c
  \]
while in category-level, primary category sales $X_{c,k_1} = \sum_{m \in \mathcal{M}_{k_1}} x_{c,m} \cdot p_{c,m}$ constrained by adjusted bounds $[L_{c,k_1}, U_{c,k_1}]$, where policy-locked SKUs fix $x_{c,m}^{\text{lock}}$ and modify bounds.

Optimization iterates per primary category $k_1$. In \textbf{High-margin material selection}, for each $k_2 \subseteq k_1$, compute margin rate $r_m = (p_{c,m} - c_{\text{cost},m}) / p_{c,m}$. Filter materials with $r_m > \text{FR}_{c,k_1} + 0.08$, rank descending, select top-5 per $k_2$ into $\mathcal{M}_{\text{cand},k_1}$. In \textbf{Sales allocation}, compute candidate sales $\hat{X}_{c,k_1}^{\text{cand}} = \sum_{m \in \mathcal{M}_{\text{cand},k_1}} \hat{s}_{c,m} \cdot p_{c,m}$:
  \begin{itemize}
    \item $\hat{X}_{c,k_1}^{\text{cand}} > U_{c,k_1}$: Scale $x_{c,m} = \hat{s}_{c,m} \cdot (U_{c,k_1} / \hat{X}_{c,k_1}^{\text{cand}})$
    \item $\hat{X}_{c,k_1}^{\text{cand}} \in [L_{c,k_1}, U_{c,k_1}]$: Set $x_{c,m} = \hat{s}_{c,m}$
    \item $\hat{X}_{c,k_1}^{\text{cand}} < L_{c,k_1}$: Fill gap $\Delta X = L_{c,k_1} - \hat{X}_{c,k_1}^{\text{cand}}$ using non-candidate materials sorted by $-r_m$, select up to 2 per $k_2$ into $\mathcal{M}_{\text{fill}}$, allocate $x_{c,m} = \min(\hat{s}_{c,m}, \Delta X / |\mathcal{M}_{\text{fill}}|)$
  \end{itemize}

Results merge across categories, yielding total recommended sales $X_c^{\text{rec}} = \sum_m x_{c,m} \cdot p_{c,m}$ and profit $\Pi_c^{\text{rec}} = \sum_m \pi_{c,m}$, ensuring optimized procurement under constraints.

\section{Experiments and Discussion}

\subsection{Comparative Analysis of Rule Generation Algorithms}
\textbf{Setup.} We evaluated the effectiveness of assortment-pricing strategies generated by different algorithms after $N=100$ iterations versus classical baselines for sales volume and profit optimization. The experimental environment used $M=10^6$ randomly sampled historical shipment records. The task required selecting a one-month assortment from $N_0=104$ SKU categories (each with $N_f=15$ features including primary/secondary categories) for $N_1=2362$ customers (each characterized by $N_c=62$ independent features), with item-level pricing recommendations. Model-generated plans were deployed under controlled pricing and fee policies to obtain actual orders. Hard constraints included: (1) total recommended purchase amounts within $\delta=25\%$ of historical averages, (2) maximum 5 SKUs per primary category, and (3) minimum 2 categories. Implementation combined market-response simulation (historical fit) and real-world deployment.

We quantitatively compared six methodologies: evolutionary algorithms, constrained reinforcement learning (RL), GPT + Monte Carlo sampling, LLM reasoning (direct rule generation), untuned LLM structure + optimizer, and fine-tuned LLM structure + optimizer. All methods underwent $N=50$ iterations. Key metrics included total sales volume, total profit, constraint-violating customer count (truncated by recommendation priority), and iterations to reach target thresholds ($B_t=6.0\times10^6$ sales, $P_t=5.0\times10^5$ profit). Experiments repeated $k=10$ times with means reported, using six mainstream LLMs detailed in our ablation study.

\begin{figure*}
  \includegraphics[width=0.95\linewidth]{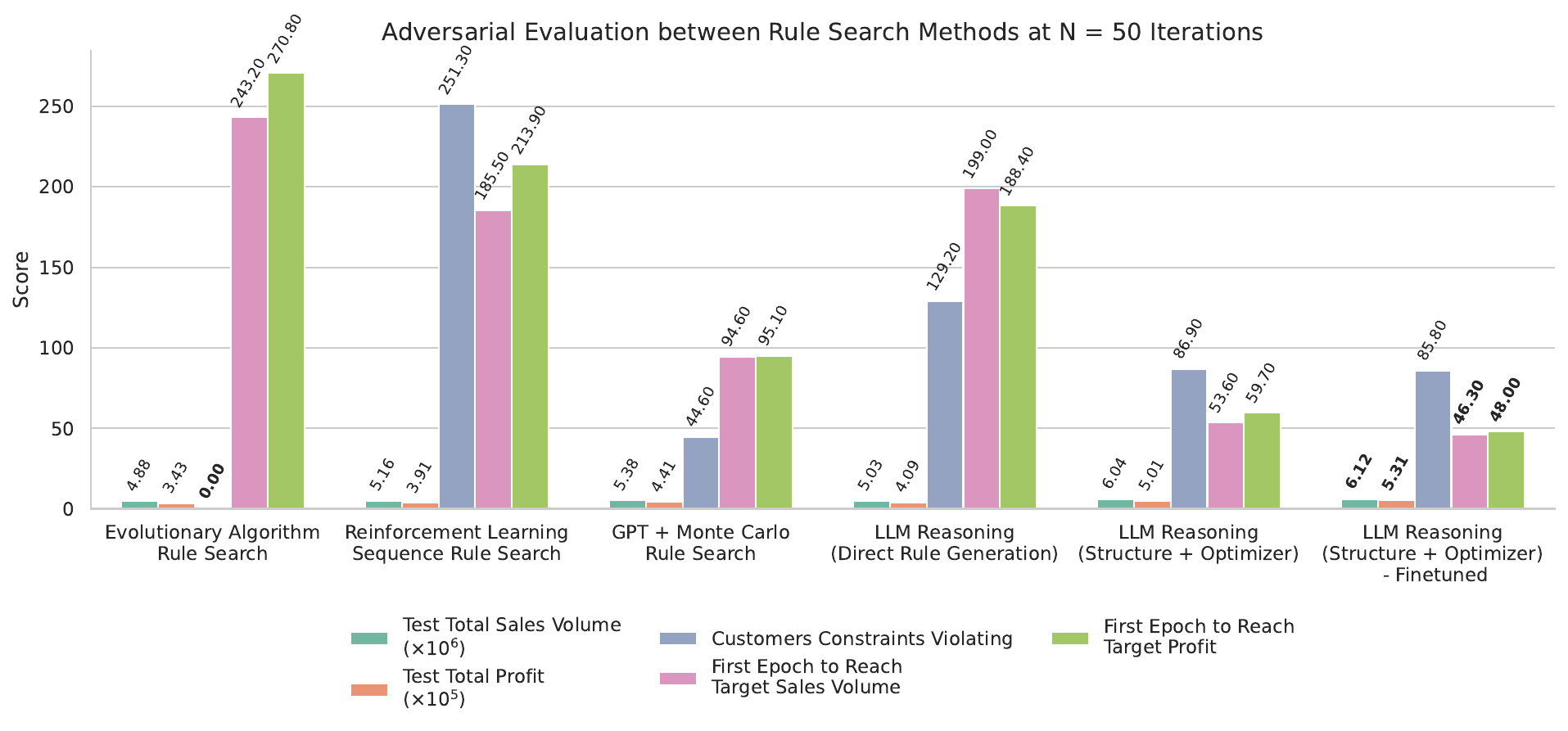}
  \caption{Adversarial Evaluation between Rule Search Methods at N=50 Iterations. }
  \label{fig:rule_search_bar}
\end{figure*}

\textbf{Results.} As shown in \ref{fig:rule_search_bar}, LLMs enhanced with trend analysis and semantic extraction demonstrated directed search capabilities, significantly improving optimization efficiency over evolutionary and RL methods. Fine-tuned LLMs showed marginal gains versus untuned versions, suggesting cost-effective deployment without fine-tuning. However, while LLMs excelled at logical structure generation (e.g. conditional statements), they exhibited weaker hyperparameter optimization and greater local oscillation tendencies than RL methods. The GPT+Monte Carlo approach outperformed classical methods but was surpassed by our LLM-based structural search, indicating modern LLMs' analytical capabilities enable more efficient solution space exploration.

\subsection{Full-Pipeline Comparison Against Baselines}

\textbf{Setup.} We evaluated end-to-end performance using an experimental environment adapted from complete historical data ($M=8.6\times10^6$ records). The task recommended assortments from $N_0=215$ SKUs (with structured tags and unstructured features like style/fragrance) to $N_1=2362$ customers. Customer features integrated three normalized, weighted sources: province/region-level historical shipment sequences ($N_{c1}=5-40$ dimensions), store demographic/geographic profiles ($N_{c2}=6-12$ dimensions), and business tags ($N_{c3}=5$ dimensions). Our framework (untuned LLM + optimizer at $N=50/100/300$ iterations) was compared against four baselines: low-rank bandit, context-based clustering, model-free pricing, and systematic B2B recommendation algorithms.

\textbf{Results.} Evaluation results \ref{tab:full_pipeline} indicate that low-rank bandit and model-free methods prioritized sales volume through aggressive pricing, achieving high sales volume but significantly lower profits with increased constraint violations. Context-based clustering sacrificed sales for higher profits, whereas the B2B recommendation algorithm achieved a balanced sales-profit tradeoff. Our LLM+optimizer framework demonstrated progressive improvement: at $N=50$ iterations, it achieved higher sales volume but slightly lower profits than clustering; at $N=100$ iterations, it surpassed clustering in both metrics while matching B2B profits with higher sales; at $N=300$ iterations, it achieved approximately 2\% further improvement, exhibiting diminishing returns relative to computational cost. These results demonstrate our framework's ability to leverage LLMs' observational and reasoning capabilities for consistent improvements over established baselines.

\begin{table}
\centering
\caption{Full-Pipeline Comparison Against Baselines: Sales Volume and Profit}
\label{tab:full_pipeline}
\begin{tblr}{
  row{1} = {c},
  cell{2}{2} = {r},
  cell{2}{3} = {r},
  cell{3}{2} = {r},
  cell{3}{3} = {r},
  cell{4}{2} = {r},
  cell{4}{3} = {r},
  cell{5}{2} = {r},
  cell{5}{3} = {r},
  cell{6}{2} = {r},
  cell{6}{3} = {r},
  cell{7}{2} = {r},
  cell{7}{3} = {r},
  cell{8}{2} = {r},
  cell{8}{3} = {r},
  hlines = {dashed},
  vline{1,4} = {-}{},
  hline{1-2,9} = {-}{solid},
}
Method                                & {Test Total \\Sales Volume \\($1.0 \times 10^6$)} & {Test Total Profit \\($1.0 \times 10^5$)} \\
Low-Rank Bandit                       & 7.450                                             & 4.174                                     \\
{Context-Based \\Clustering}          & 5.128                                             & 5.258                                     \\
{Model-Free \\Assortment Pricing}     & 7.160                                             & 4.882                                     \\
{Systematic B2B \\Recommendation}     & 5.978                                             & 5.420                                     \\
{LLM Reasoning + \\Optimizer (N=50)}  & 6.223                                             & 5.181                                     \\
{LLM Reasoning + \\Optimizer (N=100)} & 6.609                                             & 5.366                                     \\
{LLM Reasoning + \\Optimizer (N=300)} & 6.849                                             & 5.643                                     
\end{tblr}
\end{table}

\subsection{Ablation Study: Comparison Among LLM Base Models}
\textbf{Setup. }To investigate whether our framework's performance depends on specific LLM capabilities, we conducted ablation tests comparing six mainstream models: GPT-4o\cite{gpt4}, DeepSeek-R1\cite{deepseek}, Gemini\cite{gemini}, Qwen3-32B\cite{qwen3}, Claude-3.5, and Llama4-Scout\cite{llama4}. Each model was evaluated for sales and profit optimization at $N=50$ iterations under identical conditions. 

\begin{table}
\centering
\caption{Model Performance Comparison at N=50 Iterations}
\label{tab:model_comparison}
\begin{tblr}{
  row{1} = {c},
  cell{2}{2} = {r},
  cell{2}{3} = {r},
  cell{3}{2} = {r},
  cell{3}{3} = {r},
  cell{4}{2} = {r},
  cell{4}{3} = {r},
  cell{5}{2} = {r},
  cell{5}{3} = {r},
  cell{6}{2} = {r},
  cell{6}{3} = {r},
  cell{7}{2} = {r},
  cell{7}{3} = {r},
  hlines = {dashed},
  vline{1,4} = {-}{},
  hline{1-2,8} = {-}{solid},
}
Model        & {N=50 Test Total \\Sales Volume \\($1.0 \times 10^6$)} & {N=50 Test Total \\Profit \\($1.0 \times 10^5$)} \\
GPT4o        & 6.345                                                  & 5.426                                            \\
DeepSeek-R1  & 6.220                                                  & 5.230                                            \\
Gemini       & 6.171                                                  & 5.095                                            \\
Qwen3        & 5.960                                                  & 4.991                                            \\
Claude       & 6.186                                                  & 5.311                                            \\
Llama4-Scout & 5.992                                                  & 4.687                                            
\end{tblr}
\end{table}

\textbf{Results.} As displayed in \ref{tab:model_comparison}Results indicate sales and profit differences between models were predominantly below 5\%, with Llama4-Scout exhibiting a notable $\sim10\%$ profit gap. Crucially, all LLM variants significantly outperform non-LLM baselines, demonstrating that performance gains stem from the general analytical capabilities of current-generation LLMs rather than specialized mathematical competencies. This confirms the robustness of our framework across diverse LLM base models and its reliance on broad reasoning capabilities rather than model-specific strengths.

\section{Conclusion}
This work addresses multifaceted challenges encountered in physical retail industries by analyzing gaps between traditional analytical frameworks for assortment and pricing optimization and real-world operational scenarios. We identify three critical problems to solve and, for the first time, integrate generative methods with symbolic regression to tackle traditional market assortment and pricing challenges. We propose a rule search framework leveraging generative models and optimization solvers, theoretically deriving and experimentally validating its superiority over baseline generative approaches. Furthermore, our framework enhances key components including large language model (LLM)-based information extraction, customer persona aggregation, and standardized feature engineering, establishing an end-to-end data processing and optimization pipeline. Evaluations on a real-world dataset from a paper manufacturing company's online feedback system demonstrate significant advantages over existing assortment and pricing algorithms. From a market economy perspective, we provide a overall solution for capital-intensive decision-making processes.

We envision following possible research directions. \textit{Cross-Domain Generalization:} Validated in the physical paper industry, investigating efficacy in scenarios like auction bidding and mechanism design would reveal transferable principles. \textit{Insight-Driven Refinement:} Current refinement relies on LLMs' intuitive assessment. Incorporating quantitative business insights \cite{busch2024benchmarklargelanguagemodels} by data analytical coding can possibly enhance the process. \textit{Data-Scalable Generative Modeling:} For data-rich scenarios, developing domain-specific foundation models \cite{sun2024largelanguagemodelsenhanced} through historical data training could enable dedicated generative solutions.

\bibliography{aaai22}

\appendix

\section{Extended Background}
Assortment selection and pricing are classical problems in microeconomics and market engineering, playing a pivotal role in real-world economies. This problem decomposes into three interconnected components: First, customer behavior modeling requires extracting multi-dimensional features (e.g., price sensitivity, category preferences) from dynamic interaction sequences to construct personalized demand response models. Second, product feature decoupling necessitates structured representation of massive SKUs by separating intrinsic attributes from promotional effects, revealing inherent value and competitive relationships. Third, constrained decision optimization must maximize global revenue through customer-product matching under complex business rules (e.g., inventory/price constraints, tier-1/tier-2 category coverage requirements). Existing research has advanced diverse innovations across these dimensions, achieving significant theoretical and engineering progress.

\subsection{Customer Behavior Modeling: From Static Preferences to Dynamic Learning.} Accurate characterization of customer demand response underpins pricing decisions. Traditional parametric choice models (e.g., Multinomial Logit, MNL) face challenges of model misspecification and data sparsity. For dynamic feature learning, \citet{context-based} proposes an online clustering framework that aggregates observational data from products with similar demand patterns via generalized linear models and maximum likelihood estimation (MLE), mitigating cold-start issues for low-sales products. Its algorithmic core dynamically constructs an exploration mechanism based on price perturbation-neighborhood relationships, achieving an expected regret of $\tilde{O}(\sqrt{mT})$ (where $m$ denotes cluster count), outperforming the $\tilde{O}(\sqrt{nT})$ regret of single-product strategies. In temporal information fusion, \citet{Bayesian-updates} integrates online reviews into a multi-armed bandit framework, quantifying product quality perception's impact on pricing through Bayesian updates and revealing a closed-loop mechanism between dynamic pricing and market feedback. For high-dimensional sparse features, \citet{high-dimensions} employs a Regularized Maximum Likelihood Pricing (RMLP) strategy to capture critical decision variables, achieving $O(s_0 \log d \cdot \log T)$ regret under sparsity $s_0$ and providing theoretical guarantees for high-dimensional scenarios.

\subsection{Product Feature Decoupling: Value Separation and Structural Compression.}
The core product-level challenge lies in decoupling intrinsic value from external perturbations (e.g., promotions) while efficiently handling high-dimensional heterogeneous features. For value separation, \citet{ecommerce} designs a three-tier architecture: predicting baseline demand via ensemble models, computing individual price elasticities to generate candidate price-demand pairs, and solving for global optimal prices through linear programming. This approach explicitly separates brand competition effects from price elasticity differences, enabling collaborative optimization across massive SKUs. For high-dimensional representation compression, \citet{lowrank} proposes online Singular Value Decomposition (SVD) to dynamically learn latent low-rank product subspaces, compressing features to intrinsic rank $r$. This renders the regret of OPOK/OPOL algorithms independent of product count $n$, significantly enhancing scalability. Similarly, the M3P algorithm in \citet{multi-product} handles heterogeneity via a phased exploration-exploitation mechanism, achieving an $O(\log(Td)(\sqrt{T} + d\log T))$ regret lower bound. Collectively, these works demonstrate that latent structure discovery (e.g., low-rankness, clusterability) is pivotal for overcoming the curse of dimensionality.

\subsection{Constrained Decision Optimization: From Theoretical Bounds to Industrial Deployment.}
The final decision stage must balance revenue maximization with business constraints. Current research advances focus on algorithmic design and engineering: \\
1. \textit{Dynamic Strategy Frameworks}: For combinatorial optimization, \citet{assortmentselectionpricing} proposes the C-MNL model with a joint LCB-UCB strategy, employing Lower Confidence Bound (LCB) pricing to avoid price censoring and Upper Confidence Bound (UCB) assortment selection to promote exploration, achieving sublinear regret. Its extension by \citet{mnldemand} leverages MNL self-concordance to design optimistic algorithms, improving the dependency constant $\kappa^2$ for contextual bandits. In model-free optimization, \citet{model-free} constructs an Incentive-Compatible Polyhedron (IC polyhedron) encoding historical choices as valuation constraints, enabling data-driven robust pricing with linear computational complexity via the OP-MIP model. \\
2. \textit{Complex Constraint Handling}: For resource-constrained problems, \citet{primal-dual} introduces a Primal-Dual Dynamic Pricing (PD-DP) algorithm. Using Kullback-Leibler (KL) divergence mirror descent, it achieves sublinear convergence for both regret and constraint violation under non-stationary demand. \citet{constrained-pricing} develops piecewise linear approximations and MILP solvers to handle price and probability constraints with $O(1/K)$ error, extending the approach to dynamic programming decomposition frameworks. \\
3. \textit{Industrial-Grade Optimization Engines}: For ultra-large-scale scenarios, \citet{tricks} designs a maximum violation heuristic for cut generation, accelerating Lagrangian decomposition for multinational pricing across 2 million SKUs. \citet{model-distillation} proposes a "Prescriptive Student Tree" algorithm that distills black-box models into interpretable piecewise pricing strategies, bridging complex models and business deployment.

\subsection{Emerging Scenarios and Open Challenges.}
Novel technologies like generative AI are catalyzing new pricing problems. For multi-task competition scenarios, \citet{pricing-competition} constructs a price-performance-ratio optimization framework, deriving closed-form solutions via the Lambert W function for duopoly games and quantifying dynamic market information's value. Simultaneously, \citet{Perishability} formalizes data perishability challenges as time-varying coefficient models. Their Projected Stochastic Gradient Descent (PSGD) strategy achieves $O(\sqrt{T} + \sum_t \sqrt{t}\delta_t)$ regret under adversarial features. Regarding theoretical boundaries, \citet{towards-agnostic} and \citet{logarithmic-regret} respectively establish fundamental results: a minimax lower bound of $\tilde{\Omega}(T^{2/3})$ for weakly structured pricing, and $O(d\log T)$ logarithmic regret achievable under known noise distributions, providing foundational algorithm design guidance.

Despite significant algorithmic advancements, a systematic misalignment persists between theoretical frameworks and real-world economic complexities. Research on assortment-pricing optimization predominantly focuses on systematized enterprises, neglecting the vast traditional retail sector with incomplete digitalization. In these prevalent settings, substantial disparities exist between experimental conditions and operational environments. Limited digital infrastructure constrains idealized solution applicability, while conventional intelligent methods address isolated problems without incremental integration across the full retail decision chain. Moreover, widely adopted experience-based rule algorithms fail to adapt to personalized demands, yielding solutions that substantially deviate from optimality and exhibit reliability issues in unstructured, noisy contexts. We identify four critical gaps:

\textbf{Dynamic Feature Entanglement Challenges Modeling Para-digms.}
When pricing tens of thousands of SKUs in fashion e-commerce (e.g., \cite{ecommerce}), disentangling multidimensional attribute cross-effects (brand competition, seasonal trends) on demand is essential. While low-rank bandit methods (e.g., \cite{lowrank}) enhance scalability through feature subspace compression, their linear latent factor assumptions fail to capture abrupt elasticity shifts (e.g., nonlinear demand differences between luxury and FMCG goods). Concurrently, robust optimization studies (e.g., \cite{random-robust}) address model uncertainty via randomized strategies but neglect feature time-variation—illustrated by perishable goods' value decay dynamics (the time-varying coefficient challenge in \cite{Perishability}), causing static feature decoupling models to degrade in long-term decisions.

\textbf{Operational Infeasibility from Multi-Tier Business Constraints.}
Real-world decisions are constrained by hierarchical business rules: airlines face minimum seat sales requirements (e.g., guarantees studied in \cite{primal-dual}), while multinational retailers must comply with regional pricing regulations. Although constrained pricing research (e.g., \cite{constrained-pricing}) employs mixed-integer programming for price boundaries, its piecewise linear approximations lack flexibility in handling localized policy elasticity (e.g., nonlinear distributor rebate rules). Crucially, industrial-scale solutions (e.g., Lagrangian decomposition in \cite{tricks}) rely on historical solution reuse, inducing strategy degradation risks when promotions require frequent market-responsive adjustments.

\textbf{Interpretability Deficits Hinder Business Alignment.}
Black-box optimization models (e.g., Q-learning frameworks in \cite{q-learning}) adaptively adjust prices but suffer from decision opacity, preventing operational validation by business stakeholders. While model distillation techniques (e.g., \cite{model-distillation}) extract readable rules via decision trees, they disregard supply chain profit allocation requirements—such as manufacturer-distributor revenue sharing mechanisms (multilayer agency problems beyond duopoly games studied in \cite{pricing-competition})—resulting in low adoption for cross-organizational coordination.

\section{Related Work: Symbolic Regression}
Symbolic Regression (SR) autonomously discovers mathematical expressions from data without presupposing functional forms. Unlike conventional regression techniques (e.g., linear regression or neural networks), SR explores mathematical expression spaces—combinations of operators, variables, and constants—to generate interpretable closed-form equations that unveil underlying data patterns. Through multi-model rule generation augmentation, SR selectively iterates rules and generative models via dataset evaluation to optimize explicit rules.

\subsection{Traditional Symbolic Regression}
SR is extensively employed in scientific discovery, with generative frameworks evolving alongside advances in generative AI. Early SR relied predominantly on Genetic Programming (GP), which searches expression spaces through evolutionary operations. However, GP suffers from population diversity decay, hyperparameter sensitivity, and computational inefficiency, rendering it unsuitable for real-time analysis of high-dimensional market data \cite{sr-review}. To address these limitations, Deep Symbolic Regression (DSR) emerged. DSR formulates expression generation as a sequential decision-making problem; for instance, \cite{deep-sr} introduces Risk-Seeking Policy Gradients that prioritize optimization of the top-$K$ expression subsets within the reward distribution, significantly mitigating local optima traps under sparse rewards. Nevertheless, traditional DSR relies on RNN decoders with limited long-range dependency modeling, hindering their ability to capture complex nonlinear interactions among pricing factors (e.g., coupling effects between seasonal demand and supply chain fluctuations).

\subsection{Transformer-Based Architectures}
End-to-end Transformer architectures have recently gained prominence in SR research, leveraging self-attention mechanisms to globally model variable relationships and directly generate mathematical expression sequences. Representative works include \cite{symformer}, which pioneers encoder-decoder Transformers for SR by mapping data points to symbolic expressions, eliminating iterative search overhead and substantially enhancing complex formula discovery efficiency. A joint supervised learning framework \cite{transformer-plan} addresses SR's ill-posed nature arising from symbol-coefficient coupling. It proposes dual-objective optimization: (1) a weighted supervised loss for data fitting, and (2) a contrastive learning objective to decouple symbol and coefficient spaces, achieving accuracy gains in scenarios with identical expression skeletons but varying coefficients.

\subsection{Large Language Model Integration}
Large Language Models (LLMs) introduce new SR paradigms by fusing prior knowledge with few-shot reasoning. In-Context Learning (ICL) \cite{sr-llm-discovery} demonstrates that LLMs (e.g., GPT-4) infer numerical mapping rules from few-shot input-output examples, outperforming traditional models like SVM on the Friedman dataset. This capability benefits Small and Medium-sized Enterprises (SMEs) with scarce historical data by reducing cold-start costs. To address LLMs' mathematical reasoning deficiencies, \cite{gpt-guided} proposes GPT-guided Monte Carlo Tree Search (MCTS), implementing directed expression space exploration via four-phase iteration (Selection-Self-Optimization-Evaluation-Backpropagation). On the Math Odyssey dataset, LLaMA-3 8B with MCTS achieves $\sim$15\% higher accuracy than pure GPT-4, validating hybrid "LLM candidate generation + search algorithm refinement" architectures for complex pricing modeling.

\subsection{Domain-Guided SR for Economic Constraints}
Building on this foundation, we integrate SR with physical economy business rules for interpretable assortment and pricing optimization. Historically, physical economies rely on domain-expert-derived rules for stable operations. We posit that latent business rules exist within commercial transactions and warrant systematic discovery. Economic pricing must adhere to domain constraints (e.g., the law of diminishing marginal returns), necessitating domain-guided SR. Existing research offers insights: \cite{sr-physical} proposes the $\Phi$-SO framework, incorporating dimensional consistency as a hard constraint into reinforcement learning rewards to ensure physically valid expressions. Analogously, this approach extends to embedding economic priors (e.g., non-negative price elasticity, cost-revenue equilibrium) to prevent infeasible formulas. \cite{sr-review} demonstrates SR's interpretability advantage—unlike "black-box" neural networks, SR generates concise mathematical expressions that explicitly reveal functional relationships between pricing factors (e.g., traffic acquisition costs, competitive indices) and commodity prices, enabling auditing and traceability.

\section{Additional Framework Details}
\subsection{Mathematical Modeling Framework.} 
To address the partial observability of store-customer affiliations, our approach leverages large language model (LLM) technology to extract key prior knowledge from multi-source data, constructs customer feature representations, and refines them through feature engineering. The pipeline progressively advances from data parsing to feature aggregation and processing, producing efficient representations. Formally, model inputs include: customer public information (e.g., business scope and brand preferences), historical reward/penalty records, price application reports, approval documents, store-customer geolocation coordinates, demographic profiles, and SKU attributes. Outputs comprise standardized customer feature vectors $\mathbf{f}_k \in \mathbf{R}^{d_f}$ and SKU style encodings $\mathbf{e}_{\text{SKU}} \in \mathbf{R}^{d_s}$ for downstream tasks, including sales volume prediction and resource allocation.

\textbf{LLM Analysis and Prior Knowledge Extraction.} 
A locally deployed LLM parses unstructured public information to generate store affiliation priors, reward functions, and decision basis sets, addressing the challenge of implicit knowledge utilization. Specifically:  
1) For business scope, core brands, and preference texts, semantic parsing via attention mechanisms extracts key entities and relations, outputting structured prior vectors $\mathbf{p}_{\text{prior}} \in \mathbf{R}^d$ (where $d$ denotes feature dimensionality) that serve as a store affiliation knowledge base.  
2) For historical reward/penalty mechanisms, a sequence-to-sequence framework transforms text sequences $\mathbf{T} = \{t_1, t_2, \dots, t_n\}$ into reward functions $R(s, a)$. Here, $s$ represents state (e.g., customer behavior) and $a$ denotes action (e.g., approval decision), formalized as $R(s, a) = \text{MLP}(\mathbf{h}_{\text{cls}})$ where $\mathbf{h}_{\text{cls}}$ is the encoder's [CLS] token hidden state.  
3) Historical price application reports and approval documents undergo joint LLM analysis, capturing associations between numerical features (e.g., price sequences) and approval text sentiment polarity to construct decision basis sets $\mathcal{D} = \{ (\mathbf{x}_i, y_i) \}$. Here, $\mathbf{x}_i$ denotes feature vectors (e.g., price fluctuation statistics) and $y_i$ represents approval labels.  
The unified outputs---prior vector $\mathbf{p}_{\text{prior}}$, reward function $R(s,a)$, and decision set $\mathcal{D}$---provide foundational knowledge for feature aggregation.

\textbf{Multi-dimensional Demographic Profile Aggregation Strategy.} 
To address affiliation partial observability, we design a weighted aggregation mechanism based on spatial distance and operational assumptions:  
1) Spherical distance $d_{jk}$ between store $m_j$ and customer $c_k$ is computed via the Haversine formula:
\begin{equation}
d_{jk} = 2r \arcsin\left( \sqrt{\sin^2\left(\frac{\Delta \phi}{2}\right) + \cos(\phi_j) \cos(\phi_k) \sin^2\left(\frac{\Delta \lambda}{2}\right)} \right)
\end{equation}
where $\phi$ and $\lambda$ denote latitude/longitude, $r$ is Earth's radius, $\Delta \phi = \phi_j - \phi_k$, and $\Delta \lambda = \lambda_j - \lambda_k$.  
2) Customer operational radius $r_k$ correlates with business scale $s_k$ as $r_k = \alpha s_k$ ($\alpha$ estimated from historical data).  
3) Demographic features $\mathbf{g}_k$ aggregate profiles $\mathbf{q}_j$ of stores within $c_k$'s operational radius ($d_{jk} \leq r_k$). For stores covered by multiple customers, affiliation priority follows LLM-generated priors: if $\text{sim}(\mathbf{p}_{\text{prior},jk}, \mathbf{q}_j) > \theta$ (threshold $\theta$ validation-tuned), $m_j$ exclusively affiliates with $c_k$. Otherwise, distance-scale weighting applies:
\begin{equation}
w_{jk} = \frac{s_k / d_{jk}^2}{\sum_{c \in \mathcal{C}_j} s_c / d_{jc}^2}
\end{equation}
where $\mathcal{C}_j$ is the customer set covering $m_j$. Aggregated features are $\mathbf{g}_k = \sum_{j \in \mathcal{M}_k} w_{jk} \mathbf{q}_j$ ($\mathcal{M}_k$: stores covered by $c_k$).  
4) Temporal features augment through window statistics ($\mu_t$, $\sigma_t^2$ over window $W$), Fourier coefficients ($\mathbf{F} = \mathcal{F}(\mathbf{x})$, retaining top-$k$ amplitude frequencies), and sequential features (LSTM hidden states $\mathbf{h}_t$).  
The final customer representation is $\mathbf{f}_k = [\mathbf{g}_k; \mathbf{stats}_k]$, where $\mathbf{stats}_k$ concatenates statistical features.

\textbf{Feature Standardization and Encoding.}
To address feature scale disparities and high-dimensional dominance:
1) We apply Z-score standardization: $x'_{i} = \frac{x_i - \mu_i}{\sigma_i}$, updating parameters when new data exceeds threshold $N_{\text{update}}$.  
2) Dual-tower encoding prevents critical features (e.g., sales volume $y$) from being overwhelmed. This comprises:  
\quad $\bullet$ High-dimensional tower: $\mathbf{e}_{\text{high}} = \text{FC}_{\text{high}}(\mathbf{f}_k \oplus \mathbf{p})$ \\  
\quad $\bullet$ Low-dimensional tower: $\mathbf{e}_{\text{low}} = \text{ReLU}(\mathbf{W}_{\text{low}} y + \mathbf{b}_{\text{low}})$ \\  
\quad $\bullet$ Weighted fusion: $\mathbf{e}_{\text{final}} = [\beta \mathbf{e}_{\text{high}}; (1-\beta) \mathbf{e}_{\text{low}}]$ \\  
where $\beta < 0.5$ is optimized via cross-validation to emphasize key features.

\subsection{DNN Prediction Modeling}
\subsubsection{SKU Feature Representation.}
To mitigate sparsity and cold-start issues in discrete SKU attributes:  
1) Style embeddings (color/pattern) bypass one-hot encoding through:  
\quad $\bullet$ Text-based: $\mathbf{e}_{\text{style}} = \text{PCA}(\text{Bert}(\mathbf{t}_{\text{style}})) \in \mathbf{R}^{d_{\text{low}}}$ \\  
\quad $\bullet$ Multimodal: Decomposition into orthogonal components (color histogram $\mathbf{c}$, pattern distribution $\mathbf{patt}$, scent vector $\mathbf{scent}$) projected as $\mathbf{e}_{\text{style}} = \mathbf{W}[\mathbf{c}; \mathbf{patt}; \mathbf{scent}] + \mathbf{b}$.  
2) Numerical features are processed as: dense attributes (e.g., weight) used directly; sparse attributes (e.g., packaging specifications) embedded.  
The unified representation $\mathbf{e}_{\text{SKU}} = [\mathbf{e}_{\text{style}}; \mathbf{e}_{\text{dense}}; \mathbf{e}_{\text{sparse}}]$ generalizes to unseen styles via orthogonal mapping.

\subsubsection{Feature-Decoupled Sales Volume Modeling.}
To eliminate scale interference from pricing variables, we establish a direct mapping from multi-dimensional features to shipment units. Define the input feature tensor $\mathbf{X} = [\mathbf{X}_{\text{cust}} \oplus \mathbf{X}_{\text{sku}} \oplus \mathbf{X}_{\text{price}} \oplus \mathbf{X}_{\text{promo}}] \in \mathbf{R}^{N \times d}$, where $N$ is sample size, $d$ is feature dimensionality, and $\oplus$ denotes concatenation. A deep neural network $f_\theta: \mathbf{R}^d \to \mathbf{R}^+$ models the conditional expectation:
\begin{equation}
\hat{y}_{\text{units},i} = f_\theta(\mathbf{x}_i) = \sigma\left( \mathbf{W}_L \cdot \phi_{L-1} \circ \cdots \circ \phi_1(\mathbf{x}_i) + \mathbf{b}_L \right)
\end{equation}
with $\phi_l(\mathbf{z}) = \text{ReLU}(\mathbf{W}_l\mathbf{z} + \mathbf{b}_l)$ as the $l$-th layer activation, and $\sigma(\cdot)$ the output Sigmoid function. Revenue is explicitly decoupled through:
\begin{equation}
\hat{s}_{\text{revenue},i} = \hat{y}_{\text{units},i} \cdot x_{\text{price},i}
\end{equation}
This architecture ensures learning of sales dynamics rather than price-scaling effects. Parameters are optimized by minimizing the regularized mean squared error:
\begin{equation}
\mathcal{L}_{\text{pred}} = \frac{1}{N} \sum_{i=1}^N \left[ \left( f_\theta(\mathbf{x}_i) - y_{\text{units},i} \right)^2 + \lambda \|\theta\|_F^2 \right]
\end{equation}
where $\lambda \|\theta\|_F^2$ controls model complexity to prevent overfitting to price-correlated features.

\subsubsection{Rule-Prior Guided Data Augmentation with Error Control.}
To address manual annotation constraints, we adopt an RLAIF-inspired approach. Given a gold-standard customer labeling rule $g: \mathcal{X} \to \{0,1\}$, its annotation noise $\delta = |g(\mathbf{x}) - y_{\text{true}}|$ follows $\delta \sim \text{subG}(\sigma^2)$. Leveraging a pretrained language model $\Phi_{\text{LLM}}$ to parse rule semantics, we generate pseudo-labels:
\begin{equation}
    \begin{split}
        \tilde{y}_j &= \Phi_{\text{LLM}} \left( \mathcal{P}_{\text{prompt}}(g, \mathbf{x}_j) \right) \\
        \mathcal{P}_{\text{prompt}} &= \text{"Determine whether }\mathbf{x}_j\\&\text{ qualifies as a premium customer under rule }g\text{"}
    \end{split}
\end{equation}
where pseudo-label bias satisfies $\mathbf{E}[|\tilde{y}_j - y_{\text{true},j}|] \leq \eta_{\text{LLM}}$ with $\eta_{\text{LLM}} = k \cdot \exp(-I(g)/\tau)$, $I(g)$ being rule information entropy and $\tau$ a temperature coefficient. For $\mathcal{D}_{\text{aug}} = \mathcal{D}_{\text{gold}} \cup \mathcal{D}_{\text{pseudo}}$, the generalization error of $f_\theta$ admits the VC-bound:
\[
R_{\text{true}}(f_\theta) \leq \widehat{R}_{\text{aug}}(f_\theta) + \mathcal{C}(f_\theta, \delta) + \underbrace{\frac{|\mathcal{D}_{\text{gold}}|}{n}\sigma + \frac{|\mathcal{D}_{\text{pseudo}}|}{n}\eta_{\text{LLM}}}_{\Gamma_{\text{noise}}}
\]
with $n = |\mathcal{D}_{\text{aug}}|$, $\widehat{R}_{\text{aug}}$ the empirical risk, and 
\[\mathcal{C}(f_\theta, \delta) = \sqrt{\frac{\text{VC}(f_\theta)\ln(2n) - \ln\delta}{n}}\]
the complexity penalty. The noise term $\Gamma_{\text{noise}}$ quantifies label quality degradation. When $I(g) > \tau \ln(k/\sigma)$ implies $\eta_{\text{LLM}} \leq \sigma$, we obtain:
\[
\Gamma_{\text{noise}} \leq \sigma \quad \text{and} \quad \mathcal{C}(f_\theta, \delta) = O\left( \sqrt{\text{VC}(f_\theta)/n} \right)
\]
This demonstrates: 1) Label noise remains at human-annotation level ($\sigma$); 2) Increased $n$ reduces complexity penalty; 3) Pseudo-labeling achieves comparable error control when $\frac{\text{VC}(f_\theta)}{n} \ll \sigma^2$.

\subsubsection{Multi-Source Weighted Fusion for Prediction Optimization.}
The price-shipment rule base $\mathcal{R} = \{\gamma_k\}_{k=1}^K$ is partitioned into strict constraints $\Gamma_{\text{strict}}$ and soft recommendations $\Gamma_{\text{soft}}$. For each sample $\mathbf{x}_t$, we perform:

Step 1: Rule Retrieval and Weight Generation.
Relevant rules are retrieved via RAG:
\begin{equation}
\gamma_t^* = \underset{\gamma_k \in \mathcal{R}}{argmax} \cos(\psi_{\pi}(\mathbf{x}_t), \psi_{\pi}(\gamma_k))
\end{equation}
where $\psi_{\pi}$ denotes an LLM. Dynamic weight $\alpha_t$ and strictness flag $h_t$ are generated:
\begin{align}
\alpha_t &= \Phi_{\text{LLM}} \left( \mathcal{P}_{\text{weight}} \right), \\ 
\mathcal{P}_{\text{weight}} &= \text{"Input: $\mathbf{x}_t$, Rule: $\gamma_t^*$, output [0,1] weight"} \\
h_t &= \mathbf{I}[\gamma_t^* \in \Gamma_{\text{strict}}]
\end{align}

Step 2: Multi-Source Fusion.
The initial prediction $a_{\text{init}} = f_\theta(\mathbf{x}_t)$ is fused with rule output $g(\mathbf{x}_t)$:
\begin{equation}
a_{\text{fused}} = 
\begin{cases} 
g_{\text{strict}}(\mathbf{x}_t) & \text{if } h_t = 1 \\
\alpha_t a_{\text{init}} + (1 - \alpha_t) g_{\text{soft}}(\mathbf{x}_t) & \text{otherwise}
\end{cases}
\end{equation}

Step 3: Historical Trend Calibration.
Historical differential adjustment is applied:
\begin{align}
\Delta H_t &= \frac{1}{\tau} \sum_{i=1}^\tau \left( a_{t-i} - a_{t-i-\tau} \right) \\
a_{\text{final}} &= \beta \cdot \text{sgn}(\Delta H_t) \cdot \min(|\Delta H_t|, \kappa) + (1 - \beta) a_{\text{fused}}
\end{align}
with $\beta \in [0, 0.1]$ and differential threshold $\kappa$. Validated results update the rule base:
\begin{equation}
\mathcal{R} \leftarrow \mathcal{R} \cup \left\{ \left( \mathbf{x}_t, a_{\text{final}}, \text{LABEL}_{\Phi_{\text{LLM}}}(a_{\text{final}}) \right) \right\}
\end{equation}
where $\text{LABEL}$ categorizes predictions into rule types via LLM summarization.

\subsubsection{Business Data Posterior Cleaning and Validation. }
After obtaining predictions $\hat{y}_i = a_{\text{final}}(\mathbf{x}_i)$ on training/confidence set $D_{\text{train}} = \{(\mathbf{x}_i, y_i)\}_{i=1}^N$, we identify low-confidence samples using:
\begin{equation}
D_{\text{low-conf}} = \{ (\mathbf{x}_i, y_i) \in D_{\text{train}} \mid \mathcal{C}(\mathbf{x}_i, \hat{y}_i, y_i) < \delta \}
\end{equation}
where $\mathcal{C}(\cdot)$ is a confidence metric (e.g., prediction probability, label deviation) and $\delta$ a threshold. These samples undergo LLM-based cleaning:
\begin{equation}
(\text{judgment}_i, \text{reason}_i, \text{rule}_i) = \Phi_{\text{LLM}} \left( \mathcal{P}_{\text{clean}} \right)
\end{equation}
with prompt $\mathcal{P}_{\text{clean}} = \text{"Sample: }\mathbf{x}_i\text{, label: }y_i\text{. Assess label validity,}\\ \text{summarize issues if invalid, and generate cleaning rules"}$. The cleaning protocol follows:
\begin{itemize}
    \item Samples with $\text{judgment}_i = \text{"invalid"}$ are corrected/removed. Cleaning rules $\text{rule}_i$ update the automated labeling repository.
    \item Ambiguous cases ($\text{judgment}_i = \text{"uncertain"}$) are routed to manual analysis: 
    \begin{equation}
    D_{\text{manual}} = \{ (\mathbf{x}_i, y_i) \in D_{\text{low-conf}} \mid \text{judgment}_i = \text{"uncertain"} \}
    \end{equation}
\end{itemize}
Human-verified outcomes further refine the rule base and model, establishing a closed-loop data quality enhancement system.

\section{Symbolic Regression for Assortment-Pricing Strategy Modeling}
We employ symbolic regression to model assortment-pricing decisions via an interpretable mathematical function. For a feature vector $\boldsymbol{x} \in \mathbf{R}^n$ (encoding market dynamics, inventory costs, historical behavior, etc.), the function $f: \mathbf{R}^n \to \{0,1\} \times \mathbf{R}^+$ outputs both the assortment decision $d$ ($d=1$: stock; $d=0$: do not stock) and price $p$. The function decomposes as:
\[
(d, p) = 
\begin{cases} 
(1, g(\boldsymbol{x})) & \text{if } h(\boldsymbol{x}) > \tau \\
(0, \varnothing) & \text{otherwise}
\end{cases}
\]
where $h: \mathbf{R}^n \to \mathbf{R}$ determines assortment, $g: \mathbf{R}^n \to \mathbf{R}^+$ determines pricing, and $\tau$ is a threshold (typically $0$). Symbolic regression searches for $f^*$ within space $\mathcal{F}$ (comprising arithmetic operators, exponents, logarithms, features, and constants) via:
\[
f^* = \arg\max_{f \in \mathcal{F}} \Phi(f; \mathcal{D}_{\text{test}})
\]
where $\Phi$ evaluates performance on test set $\mathcal{D}_{\text{test}}$.

\subsection{Evolutionary Algorithm}
The evolutionary algorithm optimizes symbolic expressions through natural selection. An initial population $\mathcal{P}_0 = \{f_j^{(0)}\}_{j=1}^M$ of $M$ expression trees (leaf nodes: features/constants; internal nodes: operators) is randomly generated. At generation $t$:
\begin{enumerate}
    \item \textbf{Mutation:} Generate candidates $\mathcal{C}_t$ by mutating $\mathcal{P}_t$:
    \begin{itemize}
        \item \textit{Structural Expansion} ($\alpha$): Insert random subtree (depth $\delta \sim \text{Poisson}(\lambda)$) at random node
        \item \textit{Structural Pruning} ($\beta$): Replace random subtree with leaf node
        \item \textit{Parameter Perturbation} ($\gamma$): Adjust constant $\theta_k \leftarrow \theta_k + \epsilon$, $\epsilon \sim \mathcal{N}(0, \sigma^2)$
    \end{itemize}
    
    \item \textbf{Fitness Evaluation:} Evaluate $\Phi(f)$ for $\mathcal{C}_t \cup \mathcal{P}_t$:
    \begin{equation}
    \begin{split}
    \Phi(f) &= \frac{1}{|\mathcal{D}_{\text{test}}|} \sum_{i=1}^{N} \left[ \mathbf{I}_{d_i=1} \cdot \left( (p_i - c_i) \cdot q_i(p_i) \right) - \eta \cdot (p_i - p_i^*)^2 \right] \\
    &\quad - \lambda \cdot \ell(f)
    \end{split}
    \end{equation}
    Term 1 computes profit for stocked items ($c_i$: cost; $q_i(p_i)$: demand predicted via historical elasticity). Term 2 penalizes deviation from market price $p_i^*$. Term 3 penalizes complexity $\ell(f)$, defined as the node-weighted sum (operator nodes: weight 2; others: weight 1).
    
    \item \textbf{Environmental Selection:} Merge $\mathcal{P}_t$ and $\mathcal{C}_t$, rank by $\Phi(f)$, and select top $M$ individuals for $\mathcal{P}_{t+1}$.
\end{enumerate}
Iterate until fitness converges or generation limits are reached.

\subsection{Reinforcement Learning Framework}
We model symbolic expression generation as a Markov Decision Process (MDP). The state space $\mathcal{S}$ comprises partially constructed expression trees, while actions $\mathcal{A}$ involve node expansion (adding operators, features, or constants) or modification. State transitions are deterministic: $s_{t+1} = \mathcal{T}(s_t, a_t)$ yields the updated tree after applying action $a_t$. The reward function is sparse:
\[
r(s_t, a_t) = 
\begin{cases} 
\Phi(f_{s_T}) & \text{if terminal state} \\
0 & \text{otherwise}
\end{cases}
\]
where $f_{s_T}$ represents the completed function. We optimize a policy network $\pi_\phi(a|s)$ via off-policy Q-learning. Experience trajectories $\tau = (s_0,a_0,r_0,\dots,s_T)$ collected under a behavioral policy (e.g., $\epsilon$-greedy) populate replay buffer $\mathcal{B}$. The Q-function $Q_\psi(s,a)$ updates by minimizing temporal difference error:
\[
\mathcal{L}(\psi) = \mathbf{E}_{(s_t,a_t,r_t,s_{t+1}) \sim \mathcal{B}} \left[ \left( Q_\psi(s_t,a_t) - \hat{Q}(s_t,a_t) \right)^2 \right]
\]
with target values computed via target network $\bar{Q}_{\bar{\psi}}$:
\[
\hat{Q}(s_t,a_t) = r_t + \gamma \max_{a'} \bar{Q}_{\bar{\psi}}(s_{t+1}, a')
\]
Policy parameters $\phi$ update through gradient ascent:
\[
\nabla_\phi J(\phi) = \mathbf{E}_{s_t \sim \mathcal{B}} \left[ \nabla_\phi \mathbf{E}_{a_t \sim \pi_\phi(\cdot|s_t)} Q_\psi(s_t,a_t) \right]
\]
Target parameters $\bar{\psi}$ periodically synchronize with $\psi$. The optimized policy $\pi^*$ generates expression trees via greedy action selection.

\subsection{Unified Reward Function Design}
Both algorithms employ identical reward logic $\Phi(f)$ with three components:
\begin{enumerate}
    \item \textbf{Expected Profit}: $(p_i - c_i) \cdot q_i(p_i)$, where demand $q_i(p_i) = q_{i0} \cdot \exp(-\beta (p_i - p_{i0}))$ uses historical demand $q_{i0}$, price $p_{i0}$, and elasticity coefficient $\beta$
    \item \textbf{Price Deviation Penalty}: $\eta \cdot (p_i - p_i^*)^2$ ensures competitiveness
    \item \textbf{Model Complexity Penalty}: $\lambda \cdot \ell(f)$ prevents overfitting via weighted node count
\end{enumerate}
Coefficients $\eta$ and $\lambda$ balance short-term gains against long-term stability, determined via cross-validation.

\subsection{Training Mechanism for Decision LLM} 
SR search methods with LLMs are introduced in main paper, so here provide some training Formulation for specific decision LLM. 

Given sufficient historical data, we construct a parametric decision model $g_\phi: \mathcal{X} \to \mathcal{F}$ that outputs symbolic expressions for $f$, where $\mathcal{F}$ is a restricted function space (e.g., piecewise linear functions, logical expressions). Training integrates dual supervision:

\textbf{Supervised Fine-Tuning (SFT):}
Learn from expert-annotated data $\mathcal{D}_{\text{gold}} = \{(x_i, f_i^*)\}_{i=1}^N$:
\[
\mathcal{L}_{\text{SFT}} = -\mathbf{E}_{(x,f^*)\sim \mathcal{D}_{\text{gold}}} \left[ \sum_{k=1}^{|f|} \log P_\phi (f_k^* \mid f_{<k}^*, x) \right]
\]
where $f_k$ denotes the $k$-th symbol in the expression syntax tree.

\textbf{Preference Alignment:}
Utilize human-corrected data $\mathcal{D}_{\text{pref}} = \{(x, f_w, f_l)\}$ ($f_w$: corrected plan; $f_l$: original plan) via Direct Preference Optimization (DPO):
\[
\mathcal{L}_{\text{DPO}} = -\mathbf{E}_{(x,f_w,f_l)\sim \mathcal{D}_{\text{pref}}} \log \sigma \left( \beta \log \frac{P_\phi(f_w|x)}{P_{\text{ref}}(f_w|x)} - \beta \log \frac{P_\phi(f_l|x)}{P_{\text{ref}}(f_l|x)} \right)
\]
Here $P_{\text{ref}}$ is the initial model distribution and $\beta$ a temperature coefficient. This enables absorption of explicit rules and implicit preferences.

\subsubsection{Decoupled Function Structure Generation and Parameter Optimization}
To address combinatorial explosion, we propose hierarchical optimization. For function Structure $f$ with $k$ parameters $\theta \in \Theta = \prod_{i=1}^k [m_i, n_i]$, discretization step $\delta$ yields intractable search space $\prod_{i=1}^k \lfloor (n_i-m_i)/\delta \rfloor$.

\begin{itemize}
    \item \textbf{Phase 1: Function Structure Generation} \\
    Generate constrained structures via LLM:
    \[
    f(x) = \sum_{j=1}^J \alpha_j h_j(x) + \sum_{p=1}^P \beta_p \cdot \mathbf{I}(x \in \mathcal{R}_p)
    \]
    Basis functions $h_j \in \mathcal{H}$ are drawn from predefined set $\mathcal{H} = \{\text{linear terms}, \text{cross terms}, \text{sigmoid}, \text{threshold functions}\}$, $\mathbf{I}(\cdot)$ is the indicator, and $\mathcal{R}_p$ feature space regions.
    
    \item \textbf{Phase 2: Parameter Optimization} \\
    Fix Structure and solve:
    \[
    \theta^* = \underset{\theta \in \Theta}{\text{argmin}} \ L(f_\theta)
    \]
    using Sequential Quadratic Programming (SQP):
    \[
    \theta_{t+1} = \theta_t - \eta_t H_t^{-1} g_t
    \]
    where $g_t = \nabla_\theta L(f_{\theta_t})$ and $H_t$ is the Hessian approximated via BFGS:
    \[
    H_t = H_{t-1} + \frac{y_t y_t^T}{y_t^T s_t} - \frac{H_{t-1} s_t s_t^T H_{t-1}}{s_t^T H_{t-1} s_t}, \quad s_t = \theta_t - \theta_{t-1}, \quad y_t = g_t - g_{t-1}
    \]
    If $L(f_{\theta^*}) > \xi$, reconstruct framework:
    \[
    f' = \text{LLM}\left( \mathcal{P}_{\text{restruct}} \oplus \text{Tree}(f) \oplus \nabla_\theta L(f_{\theta^*}) \right)
    \]
    where $\text{Tree}(f)$ is the linearized syntax tree.
\end{itemize}
This two-stage loop reduces search complexity by decoupling combinatorial optimization into structure search and continuous parameter optimization.

\section{Example of Hierarchical Optimization Algorithm Rules}
Integrating the predictive model with our rule search framework yields novel algorithmic rules demonstrating superior simulated performance. The optimized decision function from pricing rule search is semantically summarized as a comprehensive assortment-pricing strategy, generating procurement plans through a four-stage pipeline:
\begin{enumerate}
    \item Data preprocessing
    \item Sales forecasting
    \item Cost and fee-ratio calculation
    \item Constrained optimization
\end{enumerate}
The pipeline begins by cleansing and fusing multi-source data (sales, materials, customer, cost) for predictive modeling. Subsequent stages forecast sales, quantify unit costs and category-level fee ratios, then allocate material sales volumes under business constraints to maximize profit. We formalize the modeling process with notation, formulations, and computational logic.

\subsection{Data Preprocessing}
This stage constructs a unified input dataset through customer filtering, metric computation, and data integration. Define customer set $\mathcal{C} = \{c_1, \dots, c_N\}$ and material set $\mathcal{M} = \{m_1, \dots, m_P\}$, where $N$ and $P$ denote total customers and materials. Sales data forms matrix $\mathbf{S}$ with elements $s_{c,m}$ representing historical sales volume of material $m$ to customer $c$. Price matrix $\mathbf{P}$ contains unit prices $p_{c,m}$, while forecasted sales $\hat{\mathbf{S}}$ provides $\hat{s}_{c}$ (predicted total sales for customer $c$). Candidate customers are filtered via:
\[
\mathcal{C}_{\text{cand}} = \{ c \mid \hat{s}_{c} > \theta \}, \quad \theta = 1.0
\]
targeting high-potential customers to reduce computational overhead.

Unit prices $p_{c,m}$ are extracted directly or computed as sales amount divided by volume. Material data $\mathbf{M}$ includes attributes (material codes, primary category $k_1$, secondary category $k_2$). Customer data $\mathbf{D}_{\text{cust}}$ contains identifiers and names. Data fusion via inner joins produces:
\[
\mathbf{D}_{\text{base}} = \text{sales} \bowtie \text{materials} \bowtie \text{customers}
\]
Weighted unit prices $\bar{p}_m$ reflect market averages:
\[
\bar{p}_m = \frac{\sum_{c \in \mathcal{C}} s_{c,m} \cdot p_{c,m}}{\sum_{c \in \mathcal{C}} s_{c,m}}
\]
Cost data $\mathbf{C}_{\text{cost}}$ (delivery client names, unit costs) matches customer codes via string similarity (edit distance), yielding matched matrix $\mathbf{C}_{\text{matched}}$ with elements $c_{\text{cost},m}$. The preprocessed output is:
\[
\mathbf{D}_{\text{prep}} = \mathbf{D}_{\text{base}} \cup \{\bar{p}_m\} \cup \mathbf{C}_{\text{matched}}
\]

\subsection{Sales Forecasting and Data Augmentation}
This stage generates future sales predictions using enhanced data and ML models. For each $c \in \mathcal{C}_{\text{cand}}$:
\begin{enumerate}
    \item \textbf{Expand feature space}: Form customer-material pairs $\mathcal{P}_{c} = \{ (c, m) \mid m \in \mathcal{M} \}$ via Cartesian product
    \item \textbf{Simulate discount scenarios}: Apply discrete discount levels $d \in \mathcal{D} = \{d_1, d_2, \ldots, d_Q\}$ (e.g., $\{0.9, 0.95, 1.0\}$), constructing augmented dataset $\mathbf{D}_{\text{aug}}$. Each record has feature vector $\mathbf{x}_{c,m,d} = [\text{customer attributes}, \text{material attributes}, \bar{p}_m \cdot d, \ldots]$ and target $s_{c,m}$
\end{enumerate}
Dataset size becomes $|\mathcal{C}_{\text{cand}}| \times |\mathcal{M}| \times |\mathcal{D}|$.  

A supervised model uses z-score normalized features ($\mathbf{x}_{\text{norm}} = (\mathbf{x} - \mu_{\mathbf{x}}) / \sigma_{\mathbf{x}}$) and pretrained $f_{\text{model}}$ to predict sales:
\[
\hat{s}_{c,m,d} = f_{\text{model}}(\mathbf{x}_{\text{norm}})
\]
Baseline predictions $\hat{s}_{c,m} = \hat{s}_{c,m,d=1.0}$ (no discount) feed downstream optimization.

\subsection{Cost and Fee Ratio Calculation}
This phase computes cost metrics and category-level fee ratios for economic constraints. Unit costs $c_{\text{cost},m}$ come from $\mathbf{C}_{\text{matched}}$. Fee ratios (expense-to-revenue) are allocated hierarchically. 

For primary categories $\mathcal{K}_1 = \{k_1^{(1)}, \ldots, k_1^{(R)}\}$ and secondary $\mathcal{K}_2 = \{k_2^{(1)}, \ldots, k_2^{(S)}\}$:  
\begin{enumerate}
    \item \textbf{Gross Revenue (GR)}: Historical $\text{GR}_{c,m} = s_{c,m} \cdot p_{c,m}$; category level $\text{GR}_{c,k_1} = \sum_{m \in \mathcal{M}_{k_1}} \text{GR}_{c,m}$
    \item \textbf{Category proportion}: $h_{c,k_1} = \text{GR}_{c,k_1} / \sum_{k_1'} \text{GR}_{c,k_1'}$
    \item \textbf{Fee ratios}: For expense types $\mathcal{F} = \{f_1, \ldots, f_T\}$, 
    \begin{equation}
        \begin{split}
            \text{FR}_{c,k_1,f_t} &= h_{c,k_1} \cdot \text{TotalExpense}_{f_t} / \text{GR}_{c,k_1}, \\
            \text{FR}_{c,k_1} &= \sum_{f_t \in \mathcal{F}} h_{c,k_1} \text{FR}_{c,k_1,f_t}
        \end{split}
    \end{equation}
\end{enumerate}
Monthly averaged GR $\overline{\text{GR}}_c$ constrains optimization. Final prediction data $\mathbf{D}_{\text{pred}}$ merges $\hat{s}_{c,m}$, categories, $c_{\text{cost},m}$, and $\text{FR}_{c,k_1}$.

\subsection{Constrained Optimization Decision}
This stage allocates recommended sales volumes under historical volatility and category distribution constraints to maximize profit. Define decision variable $x_{c,m}$ as recommended sales amount for material $m$ to customer $c$, with profit $\pi_{c,m} = x_{c,m} \cdot (p_{c,m} - c_{\text{cost},m})$. Optimization operates under multi-tier constraints:
In global level, Historical average order value $A_c$ bounds total recommended sales $X_c = \sum_m x_{c,m} \cdot p_{c,m}$:
  \[
  0.95 A_c \leq X_c \leq 1.05 A_c
  \]
In Category-level: Primary category sales $X_{c,k_1} = \sum_{m \in \mathcal{M}_{k_1}} x_{c,m} \cdot p_{c,m}$ constrained by:
  \[
  L_{c,k_1} \leq X_{c,k_1} \leq U_{c,k_1}, \quad
  \begin{cases} 
  L_{c,k_1} = 0.95 \cdot (h_{c,k_1} - 0.10) \cdot X_c \\
  U_{c,k_1} = 1.05 \cdot (h_{c,k_1} + 0.10) \cdot X_c
  \end{cases}
  \]
  Policy-locked SKUs (e.g., manually specified) fix $x_{c,m}^{\text{lock}}$, adjusting bounds to $[L_{c,k_1}^{\text{adj}}, U_{c,k_1}^{\text{adj}}]$.

Optimization iterates over each $k_1$:
1. \textbf{High-margin material selection}: For each $k_2 \subseteq k_1$, compute margin rate $r_m = (p_{c,m} - c_{\text{cost},m}) / p_{c,m}$. Filter materials where $r_m > \text{FR}_{c,k_1} + 0.08$, rank by $r_m$ descending, select top-5 per $k_2$ into candidate set $\mathcal{M}_{\text{cand},k_1}$.  
2. \textbf{Sales allocation}: Compute candidate sales $\hat{X}_{c,k_1}^{\text{cand}} = \sum_{m \in \mathcal{M}_{\text{cand},k_1}} \hat{s}_{c,m} \cdot p_{c,m}$:
  \begin{itemize}
    \item If $\hat{X}_{c,k_1}^{\text{cand}} > U_{c,k_1}$: Scale $x_{c,m} = \hat{s}_{c,m} \cdot (U_{c,k_1} / \hat{X}_{c,k_1}^{\text{cand}})$ (aligns with observed dealer capital constraints).
    \item If $\hat{X}_{c,k_1}^{\text{cand}} \in [L_{c,k_1}, U_{c,k_1}]$: Set $x_{c,m} = \hat{s}_{c,m}$.
    \item If $\hat{X}_{c,k_1}^{\text{cand}} < L_{c,k_1}$: Fill gap $\Delta X = L_{c,k_1} - \hat{X}_{c,k_1}^{\text{cand}}$ by selecting non-candidate materials ($\mathcal{M}_{k_1} \setminus \mathcal{M}_{\text{cand},k_1}$) sorted by loss rate $l_m = -r_m$ ascending. Add up to 2 materials per $k_2$ to filler set $\mathcal{M}_{\text{fill}}$ until coverage $\geq \Delta X$. Allocate $x_{c,m} = \min(\hat{s}_{c,m}, \Delta X / |\mathcal{M}_{\text{fill}}|)$ uniformly.
  \end{itemize}

Results are merged across categories (retaining first-occurrence for duplicates), yielding total recommended sales $X_c^{\text{rec}} = \sum_m x_{c,m} \cdot p_{c,m}$ and profit $\Pi_c^{\text{rec}} = \sum_m \pi_{c,m}$. This ensures economically optimized procurement plans under operational constraints.

\section{Impact of Generative Assortment and Pricing on Supply Chain}
Adaptive assortment selection and pricing strategies improve supply chain performance by enabling real-time customer customization, enhancing manufacturer profits, sales, and social welfare across all tiers (F to B, B2B, B2C). This approach extends game-theoretic principles to real-world economics by adapting to market transparency and dynamic conditions. In the FBBC structure, store B2's assortment choices critically influence sales, pricing, and total profit. Traditional methods (e.g., cluster-based or fixed-policy recommendations) rely on historical data or rigid rules, ignoring individual differences and market dynamics, leading to suboptimal decisions. 

Intelligent configuration enables decentralized decision-makers (particularly B2) to optimize assortments, display efforts, and prices in real-time, maximizing total supply chain profit \(\Pi_{\text{total}}\):
\[
\Pi_{\text{total}} = \sum_{i} (p_{c,i} - c_{f,i}) \times q_i
\]
where \(p_{c,i}\) is B2's terminal price, \(c_{f,i}\) is manufacturer F's production cost, and \(q_i\) is demand-driven sales. Demand \(q_i\) depends on assortment selection \(s_i\) (binary inclusion), display allocation \(a_i\) (continuous effort), price \(p_{c,i}\), regional economic indicators \(G_j\), and competitor anchor prices \(p_{\text{anchor},i}\). Constraints include information delay (\(q_i(t) = D_i(p_{c,i}(t-1), a_i, s_i, G_j)\)), channel conflict (\(|p_{c,i} - p_{\text{anchor},i}| \leq \delta_i\)), and asymmetric power (F controlling B1 via wholesale prices \(w_i\)). B2's utility balances sales velocity and markup:
\[
U_{B2} = \theta \sum_{i} q_i + (1 - \theta) \sum_{i} (p_{c,i} - p_{b1,i}) q_i
\]
with \(\theta \in [0,1]\) weighting sales volume versus profit, subject to store capacity \(\sum s_i \leq S_{\max}\) and display limits \(\sum a_i \leq A_{\max}\). Traditional methods fix \(s_i\) based on historical \(\bar{G}\), causing profit loss from adaptability gaps and double marginalization. Demand is modeled as:
\[
q_i = s_i a_i^{\gamma} d_i(p_{c,i}, G_j), \quad d_i(p_{c,i}, G_j) = \alpha_i G_j - \beta_i p_{c,i}
\]
where \(\gamma \in (0,1)\) is display elasticity, and \(\alpha_i, \beta_i\) are product-specific. When $G_j$ deviates from historical averages (variance $\sigma_G^2$), traditional methods incur error probability $p_{\text{error}} > 0$, reducing expected profit:
\begin{equation}
\begin{split}
    E[\Pi_{\text{total}}^{\text{traditional}}] &= \sum_{j} \left[ (1 - p_{\text{error}}) \Pi_{\text{total}}^{\text{optimal}}(G_j) + p_{\text{error}} \Pi_{\text{total}}^{\text{sub}}(G_j) \right]\\
    \Pi_{\text{total}}^{\text{sub}} &< \Pi_{\text{total}}^{\text{optimal}}
\end{split}
\end{equation}

Intelligent methods optimize $s_i$, $a_i$, and $p_{c,i}$ to maximize $U_{B2}$ under constraints. First-order conditions yield the optimal unconstrained price:
\[
p_{c,i}^* = \frac{\alpha_i G_j}{2\beta_i} + \frac{p_{b1,i}}{2} + \frac{\theta}{2(1 - \theta) \beta_i}
\]
where $\theta \to 1$ favors lower prices for higher sales velocity, and $\theta \to 0$ prioritizes markup. Display allocation $a_i$ is optimized via elasticity $\gamma$, and assortments $s_i$ are selected by comparing $U_{B2}$ values. This adaptation eliminates errors ($p_{\text{error}} = 0$) and aligns B2's incentives ($\theta > 0$) with supply chain goals, mitigating issues such as channel conflict (modeled via $d_i = \alpha_i G_j - \beta_i p_{c,i} - \eta (p_{c,i} - p_{\text{anchor},i})$). Profit gains under uncertainty ($G_j \sim \mathcal{N}(\mu_G, \sigma_G^2)$) are quantified as:
\[
\Delta \Pi = \int \Pi_{\text{total}}(G) (f_{\text{actual}}(G) - f_{\text{cluster}}(G))  dG > 0
\]
where gains increase with $\sigma_G^2$ and conflict intensity, amplified by real-time data and manufacturer coordination. Thus, $\Delta \Pi = f(\sigma_G^2, \theta, \text{conflict intensity}) > 0$ demonstrates the superiority of adaptive, decentralized decision-making in dynamic markets.

We envision several promising research directions to extend and refine this framework:
\begin{enumerate}
    \item \textit{Cross-Domain Generalization:} Our generative model coupled with search strategies was initially validated in experimental mathematics and physics for problems with explicit "gold-standard" rules. We have demonstrated its feasibility in the physical paper industry. Investigating its efficacy in other real-world scenarios—such as auction bidding rules, mechanism design search, and consensus protocol optimization—would reveal transferable principles and domain-agnostic capabilities.
    
    \item \textit{Insight-Driven Refinement:} The current refinement phase relies solely on LLMs' intuitive assessment of evaluation results, omitting integration of traditional business data analysis ("actionable insights") for data augmentation. Recent studies propose workflows where analytical agents generate tool token and code to do data analysis for post-evaluation \cite{wang2025toolgenunifiedtoolretrieval}. Future iterations could incorporate quantitative strategy assessments with these analytical insights to enhance the refinement process.
    
    \item \textit{Data-Scalable Generative Modeling:} Limited by sparse business data, our current approach relies on constrained fine-tuning. For data-rich scenarios, we plan to finetune problem-specific foundation models \cite{chen2025overviewdomainspecificfoundationmodel, deng2025onerecunifyingretrieverank, wang2025toolgenunifiedtoolretrieval} by tokenizing key terms in assortment and pricing problems. This transforms personalized recommendation generation for segmented customer features into a structured Q\&A paradigm, enabling training on historical annotated data to create dedicated generative models for assortment and pricing.
\end{enumerate}

\section{Parameter Configuration of Reference Algorithms}

\subsection{Evolutionary Algorithm Configuration}
The evolutionary symbolic regression implementation employs expression trees with protected operators ($\mathcal{O} = \{+, -, \times, \div, \texttt{**}, \log, \exp\}$) to evolve interpretable sales volume prediction models. Initialized with a population of 50 random trees (depth 2-5, 70\% feature leaves), the algorithm evaluates fitness via $\phi = -\text{MAE} - 0.01C - 0.005d$ where $C$ is node count and $d$ is tree depth. Genetic operations include five mutation types (30\% rate) and subtree crossover, with elitism preserving the top 10\% of solutions. Evolution proceeds for 500 generations using tournament selection and complexity-penalized fitness to balance accuracy and model interpretability, with convergence monitored via TensorBoard logging of MAE/MSE metrics.

\subsection{Reinforcement Learning Algorithm Configuration}
The reinforcement learning framework evolves symbolic regression models through an expression tree environment with operators $\mathcal{O} = \{\texttt{add}, \texttt{sub}, \texttt{mul}, \texttt{div}, \texttt{feature}, \texttt{const}\}$. State representations capture 20-dimensional expression characteristics including stack depth, operator distributions, and complexity metrics. The agent employs dual DNNs (Q-network and policy network, 256-unit hidden layers) with Adam optimization ($\eta_Q=10^{-3}$, $\eta_{\pi}=10^{-4}$), $\gamma=0.95$ discounting, and $\epsilon$-greedy exploration ($\epsilon_{\text{init}}=1.0$, $\epsilon_{\text{min}}=0.1$, decay=0.995/episode). Experiences are stored in a uniqueness-preserving replay buffer (capacity=10,000) with prioritized sampling (batch size=128). Reward design combines MAE improvement over baseline ($\Delta\text{MAE}$), complexity incentives, and validity bonuses, training for 5,000 episodes with soft target network updates ($\tau=0.01$) every 20 steps. Convergence is monitored via TensorBoard logging of reward and MAE metrics.

\subsection{SR-GPT Configuration}
\noindent The SR-GPT algorithm integrates Monte Carlo Tree Search (MCTS) with large language model guidance to evolve symbolic expressions, using a symbol library $\mathcal{S} = \{+, -, \times, \div, \wedge, \sin, \cos, \exp, \sqrt, \ln, x_1..x_{12}, c\}$. Expressions are constructed through token sequences (max length=20) validated by arity checks. At each node expansion, a 14B-parameter LLM generates symbol probabilities $P(s|\text{state})$ and state value estimates $V\in[0,1]$ via constrained JSON outputs, with $\epsilon$-greedy exploration. Reward computation employs scaled NRMSE ($S_{\text{NRMSE}}$) incorporating feature omission penalties: $R = (1 + S_{\text{NRMSE}} + 0.1\sum\sigma_{\text{unused}})^{-1}$. The MCTS executes 100 simulations per step with exploration constant $c_{\text{PUCT}}=1.0$, storing trajectories in an experience replay buffer (capacity=1000) for GPT fine-tuning. Optimization runs for 10 iterations with TensorBoard logging of reward, MAE, and RMSE metrics.

\section{ Prompts Example for Signal Regression with LLM}
Below we present an example prompt structure for generating decision functions via large language models, featuring placeholders for scenario specifications, variables, and observed outcomes.

\begin{figure*}[!th]
\begin{tcolorbox}[
    colback=blue!6!white, 
    colframe=blue!8!gray, 
    title=\textbf{Prompt for Building Initial Decision Functions}, 
    fonttitle=\bfseries\large, 
    arc=4mm, 
]
\tcblower
You are now tasked with building a decision tree model to make promotion investigation based on customer and SKU data.
Please generate a Python function that takes a DataFrame row as input and returns proposed promotion investigation.
The function should be based on the following features:
- SKU features: product-category, brand, cost, profit, wholesale-price, retail-price
- Customer features: purchase-frequency, last-period-purchases, loyalty-score, credit-score, customer-type, region
Among them types of these features are discrete string: product-category, brand, customer-type, region; and types of these features are 
integers: cost, profit, wholesale-price, purchase-frequency, loyalty-score, credit-score, last-period-purchases.

The decision tree should have a reasonable depth and complexity.

Here is an example of the function format:
def promote-sales(row):
    if row['cost'] > 100:
        if row['purchase-frequency'] > 5:
            return row['last-period-purchases'] * 1.2
        else:
            return row['last-period-purchases'] * 0.8
    else:
        if row['loyalty-score'] > 70:
            return row['last-period-purchases'] * 1.5
        else:
            return row['last-period-purchases'] * 0.9
    return row['last-period-purchases']

Notice the only requirements of the function is the return should be l column calculated by given columns (you can use + - * / ** and other const values, and the calculation operation can be put between different original columns. In addition, you can add multiple if-else to the function. All the logic in the function should be added according to  what the major loss is in history results)
\end{tcolorbox}
\caption{Construction Prompt-1}
\label{fig:build-1}
\end{figure*}

\begin{figure*}[!th]
\begin{tcolorbox}[
    colback=blue!6!white, 
    colframe=blue!8!gray, 
    title=\textbf{Prompt for Reflection and Generation Iterated Decision Functions Based on Observation and Evaluation}, 
    fonttitle=\bfseries\large, 
    arc=4mm, 
]
\tcblower
You are now tasked with building a decision tree model to make promotion investigation based on customer and SKU data.
Please generate a Python function that takes a DataFrame row as input and returns the proposed promotion investigation.
The function should be based on the following features:
- SKU features: product-category, brand, cost, profit, wholesale-price, retail-price
- Customer features: purchase-frequency, last-period-purchases, loyalty-score, credit-score, customer-type, region
Among them types of these features are discrete string: product-category, brand, customer-type, region; and types of these features are 
integers: cost, profit, wholesale-price, purchase-frequency, loyalty-score, credit-score, last-period-purchases.

Based on the previous decision tree model and selling results from promotion investigation, please reflect on the following points:

1. Analyze instances with large selling gap to identify feature combinations causing deviation
2. Check for model overfitting or underfitting
3. Identify potentially overlooked important features or feature interactions
4. Evaluate the decision tree's branching logic
5. Consider adjusting the decision tree's depth or complexity

Please modify the decision tree function based on the above reflection. The modifications should include:
1. Adjusted branching conditions
2. New or removed features
3. Adjusted termination conditions
4. Improved prediction logic

After modifying, please make new promotion investigation and record the reflections of this round and the possible causes.
Here is the previous decision tree function:
{previous-model}

Here are the previous selling results with errors compared to golden policy:
{prediction-results-with-errors}

Here are the previous error metrics:
MAE: {mae:.2f}

Reflections from previous iteration:
{reflections}

Additionally, here are some similar historical decisions and reflections for reference:
{similar-memories}
\end{tcolorbox}
\caption{Construction Prompt-2}
\label{fig:build-2}
\end{figure*}

\begin{figure*}[!th]
\begin{tcolorbox}[
    colback=blue!6!white, 
    colframe=blue!8!gray, 
    title=\textbf{Prompt for phased Reflection and Summary}, 
    fonttitle=\bfseries\large, 
    arc=4mm, 
]
\tcblower

Now please summarize completed iterative process, including:
1. The initial model construction approach
2. Major improvements in each iteration
3. Analysis of the error trend
4. Final model performance evaluation
5. Key feature identification and feature importance ranking
6. Model limitations and suggestions for improvement

Please record the complete decision tree optimization process, including the reflections of each round and possible causes.

\end{tcolorbox}
\caption{Construction Prompt-3}
\label{fig:build-3}
\end{figure*}

\section{Additional Experimental Observations}

In major experiments, $N$ denotes the number of interaction rounds. As product selection and pricing decisions involve delayed business cycles with periodic buffers, they tolerate higher per-round computation time. Additionally, real-world response costs are substantial, with extended and expensive iteration cycles. We therefore propose evaluating algorithmic strategy evolution after multiple interactions as the core metric, rather than full-process computational complexity.

Incorporating background descriptions and given priors into the prompt guiding LLM decision function generation facilitates policy optimization. For example, describing the household paper sales optimization scenario while summarizing customer-tag demographics helps the model identify influential factors and incorporate relevant features into condition settings more efficiently. To ensure cross-evaluation fairness, such introductions must exclude specific business logic (e.g., demographic preferences for product categories), relying solely on conceptual descriptions that enable the LLM to leverage its foundational knowledge. This suggests that emphasizing basic knowledge in agent systems may activate deeper model comprehension—even when such knowledge is pre-existing. This observation aligns with our finding that richer field semantic information increases the advantage of LLM generative models over exploratory methods, as they better capture field correlations to make targeted rule adjustments.

Format constraints accelerate model reasoning and optimization. Similar to \citet{gpt-guided}'s approach of structuring policy spaces on Monte Carlo trees for LLM-guided search, we construct token-level decision function prompts within our generative algorithm compiler. By providing standard structures (e.g., if-else logic for customer segmentation by sales volume), domain-specific field tokens, and function keywords, we guide the LLM's design process. Additionally, we maintain a dynamically adaptive compiler lexicon that corrects keyword misspellings via similarity matching—reducing errors that, while diminishing in recent models, persist for long field names. Compared to unrestricted string generation and parsing, this constrained approach significantly improves function parsing success rates, reduces retries, and enhances policy optimization efficiency.

Preserving optimization memory promotes monotonic improvement toward better solutions. We store each iteration's unit data—pre/post-decision functions, intermediate LLM observations, and evaluation metric deltas—in a memory module. During iterations, a similarity matching mechanism retrieves the most relevant historical unit for prompt inclusion, reusing prior corrections to prevent regression. This avoids redesign oscillations or local optimization cycles caused by forgetting earlier insights. While historically challenging for large models, recent expansions in context windows and attention capacities now enable this approach.

\section{Policy Action Distance with Golden Response}
\begin{figure}
  \includegraphics[width=0.95\linewidth]{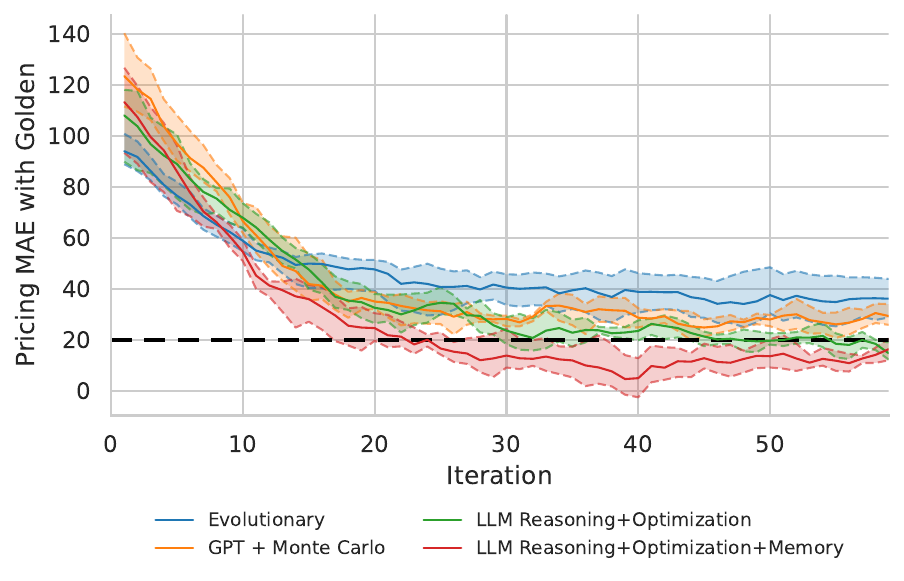}
  \caption{Adversarial Evaluation between Rule Search Methods at N=50 Iterations. }
  \label{fig:iteration}
\end{figure}
To enable real-time evaluation without physical deployment, we first construct an annotated golden response. We then visualize in Figure~\ref{fig:iteration} the evolution of MAE between policy actions discovered by different algorithms and this golden response across iterations.

We observe that evolutionary algorithms and LLM-based generative frameworks exhibit localized stepwise MAE patterns. This stems from the non-smooth optimization landscape inherent to decision rules containing conditional statements and thresholds, where validated factors persist once discovered. Upon identifying an improved policy, the algorithm archives the current function. Subsequent searches leverage this archive through pruning algorithms and memory-recall modules to constrain the search space toward optimization. Compared to evolutionary and GPT-Monte Carlo methods, Deeprule-based fitted rule generation achieves faster accuracy improvements (quicker MAE reduction). Given that structural optimizations critically enhance fitting precision, we attribute this acceleration to effective knowledge discovery via semantic associations.

Our algorithm and experiments validate symbolic regression's efficacy in deriving interpretable decision models for real economic scenarios. This applicability extends beyond product selection and pricing to other explainable rule discovery tasks, with implementation specifics being our primary focus.

As a potential extension, current LLM-based iterative mechanisms cannot guarantee monotonic optimization, occasionally exhibiting performance regression. Mitigation requires additional prior constraints or extended feedback rounds to elevate optimization lower bounds. Future work could quantify factors influencing solution quality—estimating variable impact weights, computing discovery costs per factor tier, or calculating expected improvement confidence after optimizations—to establish theoretical optimization bounds through combined metrics.

\end{document}